%% file: main.tex
\documentclass[runningheads]{llncs}

\usepackage{eccv}

\usepackage[width=122mm,left=12mm,paperwidth=146mm,height=193mm,top=12mm,paperheight=217mm]{geometry}

\usepackage{eccvabbrv}

\usepackage{graphicx}
\usepackage{booktabs}

\usepackage[accsupp]{axessibility}  %

\input{preamble}

\usepackage[pagebackref,breaklinks,colorlinks,citecolor=eccvblue]{hyperref}

\usepackage{orcidlink}

\begin{document}

\title{\ourtitle} 

\titlerunning{Tracking with Strided Memory Fusion for Consistent Vector HD Mapping}

\author{Jiacheng Chen\inst{1*} \and
Yuefan Wu\inst{1*} \and
Jiaqi Tan\inst{1*} \and Hang Ma\inst{1} \and \\ Yasutaka Furukawa\inst{1,2}}

\authorrunning{Chen et al.}

\institute{$^1$Simon Fraser University \quad $^2$Wayve}

\maketitle

\let\thefootnote\relax\footnote{*Equal contribution.
}

\input{sections/0-abs}
\input{figures/teaser}

\input{sections/1-intro}
\input{sections/2-related}
\input{sections/3-method}

\input{sections/3-evaluation}

\input{sections/4-exp}

\input{sections/5-conclusion}

\bibliographystyle{splncs04}
\bibliography{main}

\newpage
\appendix

\begin{center}
    \large \textbf{Appendix: \\ \ourtitle}
\end{center}

\noindent The appendix provides remaining system details (\S\ref{supp:method_details},\S\ref{supp:consistent_benchmark},\S\ref{supp:online_merging}) and additional experimental results (\S\ref{supp:exp}) as mentioned in the main paper. 

\begin{itemize}[leftmargin=*,itemsep=1pt]

\item[$\diamond$] \S\ref{supp:method_details}: Remaining details of \ourmethod, including 1) The transformation loss for PropMLP; 2) 
The query propagation module;
and 3) The strided memory selection mechanism.

\item[$\diamond$] \S\ref{supp:consistent_benchmark}: Remaining details/analyses of our consistent vector HD mapping benchmarks, including details on 1) consistent ground truth generation; 2) the track extraction algorithm for detection-based baseline approaches; and 3) 
the consistency-aware mAP (C-mAP).

\item[$\diamond$] \S\ref{supp:online_merging}: Details of our online merging algorithm that generates the global vector HD maps from per-frame reconstructions.

\item[$\diamond$] \S\ref{supp:exp}: Additional experimental results and analyses, including 1) The C-mAP results of all methods using the track extraction algorithm with different lookback parameters; 2) Check the temporal consistency of MapTR's ground truth data using our consistent-aware benchmarks; and 3) Additional qualitative results. 

\end{itemize}

\input{sections_supp/supp-implementations}

\input{sections_supp/supp-exp}

\end{document}

%% file: preamble.tex
\usepackage{booktabs}
\usepackage{caption}
\usepackage{multirow}
\usepackage{enumitem}
\usepackage{comment}
\usepackage{amsmath}
\usepackage{xspace}

\usepackage{tabu}

\usepackage{threeparttable}
\usepackage{arydshln}
\usepackage{comment}

\usepackage{mathtools}

\usepackage{pifont,hologo}
\newcommand{\cmark}{\text{\ding{51}}}

\newcommand{\supp}{Appendix\xspace}

\newcommand{\ourtitle}{MapTracker: Tracking with Strided Memory Fusion for Consistent Vector HD Mapping}

\newcommand{\ourmethod}{MapTracker\xspace}

\newcommand\myparafirst[1]{\vspace{0.3em}\noindent\textbf{#1.}}

\newcommand\mypara[1]{\vspace{0.3em}\noindent\textbf{#1.}}

\newcommand\myparapara[1]{\vspace{0.3em}\noindent\textbf{#1}}

\definecolor{deemph}{gray}{0.6}

\newcommand{\allImageInfo}{\mathbf{I}}

\newcommand{\pose}{P}

\newcommand{\curve}{V}

\newcommand{\vecOutputs}{\mathbf{Y}}
\newcommand{\vecOutput}{Y}

\newcommand{\bevmemInit}{\mathbf{M}_{\textrm{BEV}}^{\textrm{init}}}

\newcommand{\vecmemInit}{\mathbf{M}_{\textrm{VEC}}^\textrm{init}}

\newcommand{\membev}{\mathbf{M}_{\textrm{BEV}}}
\newcommand{\memvec}{\mathbf{M}_{\textrm{VEC}}}
\newcommand{\bevseg}{\mathbf{S}}

\newcommand{\SelectTimes}{\boldsymbol{\pi}}

\newcommand{\matching}{\boldsymbol{\omega}}
\newcommand{\matchingNew}{\boldsymbol{\omega}_\text{new}}
\newcommand{\matchingTrack}{\boldsymbol{\omega}_\text{track}}

\definecolor{ourgreen}{HTML}{19b65d}

\usepackage{algorithm}
\usepackage{algorithmic}
\usepackage{bm}

%% file: sections/0-abs.tex
\begin{abstract}
This paper presents a vector HD-mapping algorithm that formulates the mapping as a tracking task and uses a history of memory latents to ensure consistent reconstructions over time.
Our method, \textit{\ourmethod}, accumulates a sensor stream into memory buffers of two latent representations: 1) Raster latents in the bird's-eye-view (BEV) space and 2) Vector latents over the road elements (i.e., pedestrian-crossings, lane-dividers, and road-boundaries).
The approach borrows the query propagation paradigm from the tracking literature that explicitly associates tracked road elements from the previous frame to the current, 
while fusing a subset of memory latents selected with distance strides to further enhance temporal consistency.
A vector latent is decoded to reconstruct the geometry of a road element. 
The paper further makes benchmark contributions by 1) Improving processing code for existing datasets to produce consistent ground truth with temporal alignments and 2) Augmenting existing mAP metrics with consistency checks.
\ourmethod significantly outperforms existing methods on both nuScenes and Agroverse2 datasets by over 8\% and 19\% on the conventional and the new consistency-aware metrics, respectively.
The code and models are available on our project page: \url{https://map-tracker.github.io}.
\end{abstract}

%% file: figures/teaser.tex
\begin{figure}[t]
\centering
\includegraphics[width=\textwidth]{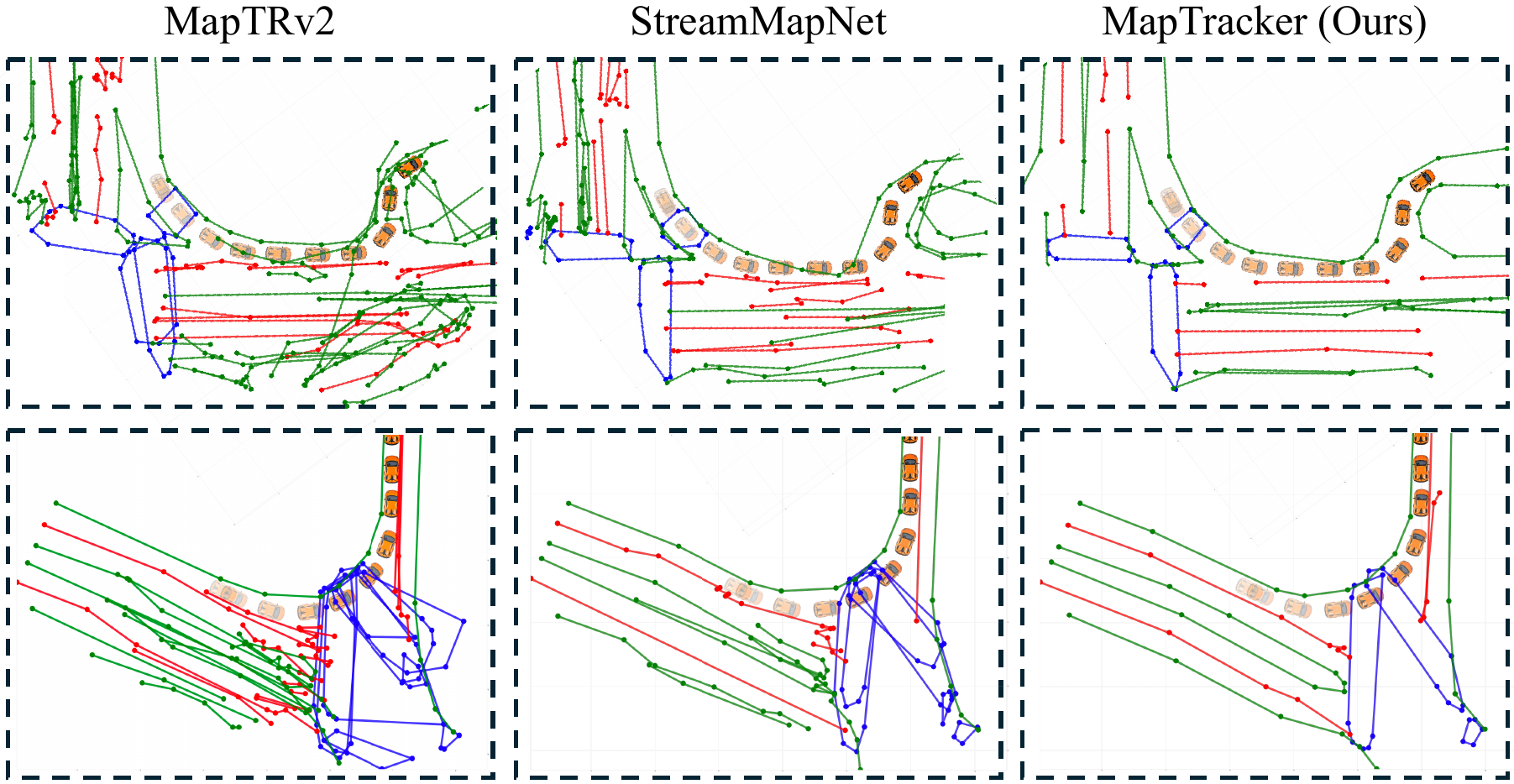}
\caption{\ourmethod produces high-quality and temporally consistent vector HD maps, which are progressively merged into a global vector HD map by a simple online algorithm. 
The current state-of-the-art methods, MapTRv2~\cite{liao2023maptrv2} and StreamMapNet~\cite{yuan2024streammapnet}, fail to produce consistent reconstructions, leading to very noisy global maps.
The figure shows two challenging scenarios (cars are turning) from the nuScenes\cite{caesar2020nuscenes} dataset.
}
\label{fig:teaser}
\end{figure}

%% file: sections/1-intro.tex
\section{Introduction}

Humans forget, so do neural networks. A robust memory is crucial for online systems to produce consistent outputs. Vector HD mapping, a task of reconstructing vector road geometries from vehicle sensor data, has made dramatic progress.
A consistent vector HD mapping system, capable of reconstructing a consistent HD map of a city from a single drive-through (See \autoref{fig:teaser} for examples), would have a tremendous impact on our society, reducing the cost of HD map creation for tens of thousands of cities in the world and enhancing the safety and stability of self-driving cars.

Existing vector HD mapping methods~\cite{yuan2024streammapnet,liao2022maptr,liao2023maptrv2,liu2023vectormapnet,li2022hdmapnet} focus on per-frame reconstruction via detection-style transformer networks~\cite{carion2020detr}. 
They detect road elements anew in every frame without consistency enforcement, potentially guided by reconstructions from the previous frame.
Furthermore, a standard recurrent latent embedding is often the choice for memory mechanism~\cite{yuan2024streammapnet}, where accumulating the entire history in a single latent memory proves challenging, especially for cluttered environments with numerous vehicles obscuring road structures.

Towards ultimate temporal consistency, this paper presents \ourmethod with two key design elements.
First, tracking instead of detection becomes the formulation, specifically borrowing the query propagation paradigm from the tracking literature that explicitly associates tracked road elements across frames. 
Second, a sequence of memory latents from past frames serves as the memory mechanism. 
Concretely, we retain memory buffers for two latent representations from the past frames: 1) Raster latents in the bird's-eye-view (BEV) space and 2) Vector latents over the tracked road elements, while using a subset of memory latents based on distance strides for effective information fusion. A vector latent reconstructs a road element geometry.

To prepare the tracking labels and measure the consistency of HD map reconstructions,
this paper introduces a new benchmark based on nuScenes~\cite{caesar2020nuscenes} and Agroverse2~\cite{wilson2023argoverse} datasets.
Specifically, we improve the processing code of the two datasets to produce consistent ground truth data with temporal alignments, then propose a consistency-aware mean average precision (mAP) metric.

We have made extensive comparative evaluations based on the traditional and the new mAP metrics. {\ourmethod} significantly outperforms the competing methods by over 8\% on the conventional distance-based mAP, reaching 76.1 mAP on nuScenes and 76.9 mAP on Argoverse2.  
With the new consistency-aware metrics, \ourmethod~demonstrates superior temporal consistency and improves the StreamMapNet baseline by over 19\%.

To summarize, this paper makes three contributions: 1) A novel vector HD mapping algorithm that formulates HD mapping as tracking and leverages the history of memory latents in two representations to achieve temporal consistency; 2) An improved vector HD mapping benchmark with temporally consistent ground truth and a consistency-aware mAP metric; and 3) SOTA performance with significant improvements over the current best methods on traditional and new metrics. The code and the new benchmark data will be available. 

%% file: sections/2-related.tex
\section{Related Work}
This paper tackles consistent vector HD mapping by 1) borrowing an idea from the visual object tracking literature and 2) devising a new memory mechanism. The section first reviews recent trends in visual object tracking with transformers and memory designs in vision-based autonomous driving. Lastly, we discuss competing vector HD mapping methods.

\mypara{Visual object tracking with transformers}
Visual object tracking~\cite{yilmaz2006object-tracking-survey} has a long history, where end-to-end transformer~\cite{vaswani2017transformer} methods become a recent trend due to the simplicity.
TrackFormer~\cite{meinhardt2022trackformer}, TransTrack~\cite{sun2020transtrack}, and MOTR~\cite{zeng2022motr,zhang2023motrv2} leverage the attention mechanism with \textit{track queries} to explicitly associate instances across frames.
MeMOT~\cite{cai2022memot} and MeMOTR~\cite{gao2023memotr} further extend the tracking transformers with memory mechanisms for better long-term consistency.
This paper formulates vector HD mapping as a tracking task by incorporating track queries with a more robust memory mechanism.

\mypara{Memory designs in autonomous driving}  
Single-frame self-driving systems have difficulty in handling occlusion, sensor failure, or complex environments. Temporal modeling with memories offers promising complements\cite{huang2022bevdet4d,yang2023bevformerv2,han2023videobev,li2023memoryseg,lin2023sparse4d-v2,lin2023sparse4d-v3,yuan2024streammapnet,wang2024sqd-mapnet,gu2023vip3d}. Many memory designs exist for the raster BEV  features~\cite{li2022bevformer,philion2020LSS}, which form the foundation of most autonomous-driving tasks~\cite{ma2022vision-bev-survey,li2023delving-bev-survey}.
BEVDet4D~\cite{huang2022bevdet4d} and
BEVFormerv2~\cite{yang2023bevformerv2} stack features of multiple past frames as a memory,
but the computation scales linearly with history length, struggling to capture long-term information. 
VideoBEV~\cite{han2023videobev} propagates BEV raster queries across frames to accumulate information recurrently. In the vector domain, Sparse4Dv2~\cite{lin2023sparse4d-v2} employs a similar RNN-style memory for object queries, while Sparse4Dv3~\cite{lin2023sparse4d-v3} further uses temporal denoising for robust temporal learning. These ideas have been partially incorporated by vector HD mapping approaches~\cite{yuan2024streammapnet,wang2024sqd-mapnet}. 
This paper proposes a new memory design for both the raster BEV latents and the vector latents of road elements.

\myparafirst{Vector HD mapping}
Traditionally HD maps are reconstructed offline with SLAM-based methods~\cite{zhang2014loam,shan2018lego-loam,shan2020lio-sam}, followed by human curation,
requiring high maintenance costs.
Online vector HD mapping algorithms are gaining more interest over their offline counterparts as their accuracy and efficiency improve, which would simplify the production pipeline and handle map changes.
HDMapNet~\cite{li2022hdmapnet} turns raster map segmentation into vector map instances via post-processing and has established the first Vector HD mapping benchmark.
VectorMapNet~\cite{liu2023vectormapnet} and MapTR~\cite{liao2022maptr} both leverage DETR-based~\cite{carion2020detr} transformers for end-to-end prediction. The former predicts the vertices of each detected curve autoregressively, while the latter uses hierarchical queries and matching loss to predict all the vertices simultaneously. MapTRv2~\cite{liao2023maptrv2} further complements MapTR with auxiliary tasks and network modifications. Curve representation~\cite{qiao2023bemapnet,ding2023pivotnet,zhang2023gemap}, network design~\cite{xu2023insightmapper}, and training paradigm~\cite{zhang2023mapvr,chen2023polydiffuse} are the focus of other works.
StreamMapNet~\cite{yuan2024streammapnet} steps towards consistent mapping by borrowing the streaming idea from BEV perception. The idea accumulates the past information into memory latents and passes as a condition (i.e., a conditional detection framework).
SQD-MapNet~\cite{wang2024sqd-mapnet} proposes temporal curve denoising to facilitate temporal learning, mimicking DN-DETR~\cite{li2022dn-detr}. 

%% file: sections/3-method.tex
\section{\ourmethod}
\label{sec:method}

\input{figures/method_overview}

A robust memory mechanism is the core of \ourmethod, accumulating a sensor stream into latent memories of two representations: 1) Bird's-eye-view (BEV) memory of 
a region around a vehicle in the top-down BEV coordinate frame as a latent image; and 2) Vector (VEC) memory of road elements (i.e., pedestrian-crossings, lane-dividers, and road-boundaries) as a set of latent vectors.

Two simple ideas with the memory mechanism achieve consistent mapping. The first idea is to use a buffer of memories from the past instead of a single memory at the current frame~\cite{yuan2024streammapnet,han2023videobev,li2022bevformer}. 
A single memory should hold the entire past information but is susceptible to memory loss, especially in cluttered environments with numerous vehicles obscuring road structures. Concretely, we select a subset of the past latent memories for fusion at each frame
based on the vehicle motions for efficiency and coverage. The second idea is to formulate the online HD mapping as a tracking task. The VEC memory mechanism maintains a sequence of memory latents with each road element and makes this formulation straightforward by borrowing a query propagation paradigm from the tracking literature.
The rest of the section explains our neural architectures (See \autoref{fig:method_overview} and \autoref{fig:equations}), consisting of the BEV and VEC memory buffers and their corresponding network modules, and then presents the training details.

\input{figures/equations}

\subsection{Memory Buffers}
A BEV memory, $\membev(t) \in\mathbb{R}^{50 \times100 \times 256}$, is a 2D latent image in the BEV coordinate frame centered and oriented with the vehicle at frame $t$. The spatial dimension (i.e., $50\times 100$) covers a rectangular area, 15m left/right and 30m front/back. 
Each memory latent accumulates the entire past information, while the buffer holds such memory latents for the last 20 frames, making the memory mechamism redundant but robust.

A VEC memory, $\memvec(t) 
 \in \{ \mathbb{R}^{512} \}$, is a set of vector latents, each of which accumulates information of an active road element up to frame $t$. 
The number of active elements 
varies per frame.
The buffer holds the latent vectors of the past 20 frames and their correspondences across frames (i.e., a sequence of vector latents corresponding to the same road element).

\subsection{BEV Module}

\sloppy  {\bf Inputs} 
are 1) CNN features of the onboard perspective images processed by the image backbone (the official ResNet50 model~\cite{he2016resnet} pretrained on ImageNet~\cite{deng2009imagenet})
and their camera parameters $\allImageInfo(t)$;
2) the BEV memory buffer $\left\{ \membev(t-1), \membev(t-2), ...\right\}$; and 3) the vehicle motions
$\left\{ \pose^t_{t-1}, \pose^t_{t-2}, ... \right\}$, where
$\pose_{t_1}^{t_2} \in \mathbb{R}^{4 \times4}$ is the affine transformation of the vehicle coordinate frame from frame $t_1$ to $t_2$. The following explains the four components of the BEV module architecture and its outputs. 

\mypara{[1. BEV Query Propagation]}
A BEV memory is a 2D latent image in a vehicle coordinate frame. An affine transformation $\pose^t_{t-1}$ and a bilinear interpolation initialize the current BEV memory $\membev(t)$ with the previous one $\membev(t-1)$. 
For pixels that fall outside the latent image after the transformation, per-pixel learnable embedding vectors $\bevmemInit \in\mathbb{R}^{50 \times100 \times 256}$ 
are the initialization instead, whose operation is denoted as ``MaskBlend'' in \autoref{fig:equations}.

\mypara{[2. Deformable Self-Attention]} A deformable self-attention layer~\cite{zhu2020deformable-detr} enriches the BEV memory $\membev(t)$.

\mypara{[3. Perspective-to-BEV Cross-Attention]} Similar to StreamMapNet~\cite{yuan2024streammapnet},
a spatial deformable cross-attention layer from BEVFormer~\cite{li2022bevformer} injects the perspective-view information $\allImageInfo(t)$ into $\membev(t)$, followed by a standard feed-forward network (FFN) layer~\cite{vaswani2017transformer}.

\mypara{[4. BEV Memory Fusion]}
The memory latents in the buffer are fused to enrich $\membev(t)$. Using all the memories is computationally expensive and redundant. We use a strided selection of four memories without repetition, whose vehicle positions are the closest to (1m/5m/10m/15m)
from the current position.
An affine transformation and a bilinear interpolation align the coordinate frames of the selected memories to the current: $\{ \membev^{*}(t'), t' \in \SelectTimes(t) \}$, where $\SelectTimes(t)$ denotes the selected times. We concatenate $\membev(t)$ with the aligned memories and use a lightweight residual block with two convolution layers to udpate $\membev(t)$.
The last three components of the BEV module repeat twice without weight sharing.

\myparapara{Outputs}
are 1) the final memory $\membev(t)$ saved to the buffer and passed to the VEC module; and 2) the rasterized road element geometries $\bevseg(t)$ which is inferred by a segmentation head and used for a loss calculation (See \S\ref{subsec:training}). The segmentation head is a linear projection module that projects each pixel in the memory latent to a 2$\times$2 segmentation mask, thus producing a 100$\times$200 mask.

\subsection{VEC Module}

\textbf{Inputs} 
are 1) the BEV memory $\membev(t)$; 2) the vector memory buffer $\left\{ \memvec(t-1), \memvec(t-2), ...\right\}$; and 3) the vehicle motions
$\left\{ \pose^t_{t-1}, \pose^t_{t-2}, ... \right\}$.

\mypara{[1. Vector Query Propagation]}
A vector memory is a set of latent vectors of the active road elements. Borrowing the query propagation paradigm from transformer-based tracking approaches~\cite{zeng2022motr,meinhardt2022trackformer,sun2020transtrack}, we initialize the vector memory as
$\memvec(t) = \left[ \memvec^{\text{prop}}(t), \memvec^{\text{new}}(t) \right]$.
$\memvec^{\text{new}}(t)$ denotes 100 latent vectors for 100 new road element candidates, which are initialized with 100 learnable embeddings $\vecmemInit$.
$\memvec^{\text{prop}}(t)$ denotes latent vectors for the currently tracked road elements, which are initialized with the corresponding latent vectors in the previous memory $\memvec(t-1)$ after using a two-layer MLP to align the coordinate frame.
Concretely, we turn $\pose^t_{t-1}$ into a 4D vector of rotation quaternion and 3D vector of translation parameters, represent with their positional encodings~\cite{vaswani2017transformer}, concatenate them with each vector latent in $\memvec(t-1)$, and apply an MLP.
We call it PropMLP, which handles the temporal propagation.

\mypara{[2. Vector Instance Self Attention]} Similar to StreamMapNet, a standard self-attention layer enriches the vector latents in the memory $\memvec(t)$.

\mypara{[3. BEV-to-Vector Cross Attention]} 
The Multi-Point Attention from StreamMapNet, which is an extension of the vanilla deformable cross-attention~\cite{zhu2020deformable-detr}, injects the BEV information from $\membev(t)$ into $\memvec(t)$.

\mypara{[4. Vector Memory Fusion]}
For each latent vector in the current memory $\memvec(t)$, latent vectors in the buffer associated with the same road element are 
fused to enrich its representation.
The same strided frame-selection chooses four latent vectors, where the selected frames $\SelectTimes(t)$ would be different and fewer for some road elements with a short tracking history.
For example, an element that has been tracked for two frames has only two latents in the buffer.
A standard cross-attention followed by an FFN layer injects the selected latents after aligning their coordinate frames by the same PropMLP module $\{ \memvec^*(t'), t' \in \SelectTimes(t)\}$. To be precise, a query is a latent in $\memvec^{\text{prop}}(t)$, where a key/value is a latent in $\{ \memvec^*(t'), t' \in \SelectTimes(t)\}$.
In \autoref{fig:equations},
we omit the element index for the ``Per-Ins-CA'' operation, which acts on each element independently.
The last three components of the VEC module repeat six times without weight sharing.

\myparapara{Outputs}
are 1) the final memory 
$\memvec(t)$ for ``positive'' road elements that pass the classification test by a single fully connected layer from $\memvec(t)$; and 2)
vector road geometries of the positive road elements, 
regressed by the 3-layer MLP from $\memvec(t)$.
The threshold of the classification test is 0.4 for the first frame, and 0.5/0.6 for the propagated/new road elements for subsequent frames.
$\vecOutputs(t) = \{ \left( \curve_i, p_i \right) \}$ denotes the outputs.
Following the prior convention~\cite{liao2022maptr}, each element geometry $\curve_i = {\left[ (x_1, y_1),\dots, (x_{20}, y_{20}) \right]}$ is a polygonal curve with 20 points in the BEV coordinate frame. $p_i$ is the class probability score.

\subsection{Training}
\label{subsec:training}
The ground-truth road element geometries are denoted as
$\hat{\vecOutputs}(t) = \{ 
\hat{\vecOutput}_i \}$. $\hat{\vecOutput}_i = (\hat{\curve}_i, \hat{c}_i)$,  where $\curve_i$ has 20 points interpolated from the raw ground-truth vector. $\hat{c}_i$ is the class label.
Standard OpenCV and PIL libraries rasterize $\hat{\vecOutputs}(t)$ on an empty BEV canvas to obtain the ground-truth segmentation image $\hat{\bevseg}(t)$

\mypara{BEV loss}
We employ per-pixel Focal loss~\cite{lin2017focal} and per-class Dice loss~\cite{milletari2016dice} on the BEV outputs $\bevseg(t)$, which are common auxiliary losses in vector HD mapping approaches~\cite{qiao2023machmap,liao2023maptrv2}. The loss is defined by
\begin{equation}
\mathcal{L}_\text{BEV} = \lambda_1 \mathcal{L}_\text{focal}(\bevseg(t), \hat{\bevseg}(t))+ \lambda_2 \mathcal{L}_\text{dice}(\bevseg(t), \hat{\bevseg}(t))
\end{equation}

\mypara{VEC loss}
Inspired by MOTR~\cite{zeng2022motr}, an end-to-end transformer for multi-object tracking, we extend the matching-based loss~\cite{liao2022maptr,yuan2024streammapnet} to explicitly consider ground-truth tracks (See \S\ref{sec:benchmark} for ground-truth processing). 
For each frame $t$,  $\hat{\vecOutputs}(t)$ consists of two disjoint subsets: new elements $\hat{\vecOutputs}_\text{new}(t)$ and tracked elements $\hat{\vecOutputs}_\text{track}(t)$. For the vector outputs, we denote the results from the propagated latents $\memvec^{\text{prop}}(t)$ as $\vecOutputs_\text{track}(t)$, and results from the new latents $\memvec^{\text{new}}(t)$ as $\vecOutputs_\text{new}(t)$. 
Note that to make the VEC module robust to potential errors in pose estimation, we randomly perturb the transformation matrix $\pose^t_{t-1}$ by adding a Gaussian noise during training. 
We train the module with a tracking loss that explicitly considers the temporal alignments. \S\ref{sec:benchmark} explains
the ground-truth preparation.
The optimal instance-level label assignment for new elements
is defined as:
\begin{equation}
\matchingNew (t) = \underset{\matchingNew(t) \in \boldsymbol{\Omega}(t) }{\mathrm{arg}\min}~\mathcal{L}_\text{match}(\hat{\vecOutputs}_\text{new}(t)|_{\matchingNew(t)}, \vecOutputs_\text{new}(t)).
\end{equation}
$\boldsymbol{\Omega}(t)$ is the space
of all bipartite matches. $\mathcal{L}_\text{match}$ is the hierarchical matching cost similar to the one proposed in MapTR\cite{liao2022maptr}, consisting of a focal loss $\mathcal{L}_\text{focal}(\{ \hat{c}_i\}|_{\matchingNew(t)}, \{ p_i \} )$ and a permutation-invariant line coordinate loss $\mathcal{L}_\text{line}(\{ \hat{\curve}_i\}|_{\matchingNew(t)}, \{ \curve_i \} )$.
The label assignments $\matching(t)$ between all outputs and ground truth is then defined inductively:
\begin{align} 
\matching(t) = \matchingTrack(t) \cup \matchingNew(t);~
\matchingTrack(t) = \begin{cases}
    \varnothing, & \text{if}~ t = 0 \\
    \matching(t-1), & \text{if}~ t > 0
\end{cases}.
\end{align} 
$\matchingTrack(t)$ is the label assignments between $\vecOutputs_\text{track}(t)$ and $\hat{\vecOutputs}_\text{track}(t)$. The tracking-style loss for the vector outputs is:
\begin{align}
\mathcal{L}_\text{track} =  \lambda_3 \mathcal{L}_\text{focal}(\{ \hat{c}_i\}|_{\matching(t)}, \{ p_i \} ) + \lambda_4 \mathcal{L}_\text{line}(\{ \hat{\curve}_i\}|_{\matching(t)}, \{ \curve_i \} ).
\end{align}

\mypara{Transformation loss} 
We borrow the transformation loss $L_\text{trans}$ from StreamMapNet~\cite{yuan2024streammapnet} to train the PropMLP, which enforces that the query transformation in the latent space maintains the vector geometry and class type. Full details are provided in the \supp. The final training loss is
\begin{equation}
\mathcal{L} = \mathcal{L}_\text{BEV} + \mathcal{L}_\text{track} + \lambda_5 \mathcal{L}_\text{trans}.
\end{equation}

\mypara{Training details}
For each training sample, we randomly choose 4 out of the previous 10 frames to compose a training clip with a length of 5. We freeze the image backbone for the first four training frames to reduce the memory cost for the clip-based training.
The training of the system has three stages:
1) Pre-train the image backbone and BEV encoder with only $\mathcal{L}_\text{BEV}$; 2) Warm up the vector decoder while freezing all other parameters with $\mathcal{L}$, where the vector memory is turned on after 500 warmup iterations; 3) Jointly train all parameters with $\mathcal{L}$. 
The second stage warms up the vector module with a large batch size to facilitate initial convergence, as we cannot afford in the joint training.
The loss weights are $\lambda_1=10.0$, $\lambda_2=1.0$, $\lambda_3=5.0$, $\lambda_4=50.0$, $\lambda_5=0.1$.
We use an AdamW~\cite{loshchilov2018fixing} optimizer with an initial learning rate 5e-4 and the weight decay is set to 0.01. A cosine learning rate scheduler is used with a final learning rate of 1.5e-6.

%% file: figures/method_overview.tex
\begin{figure}[t]
\centering
\includegraphics[width=\textwidth]{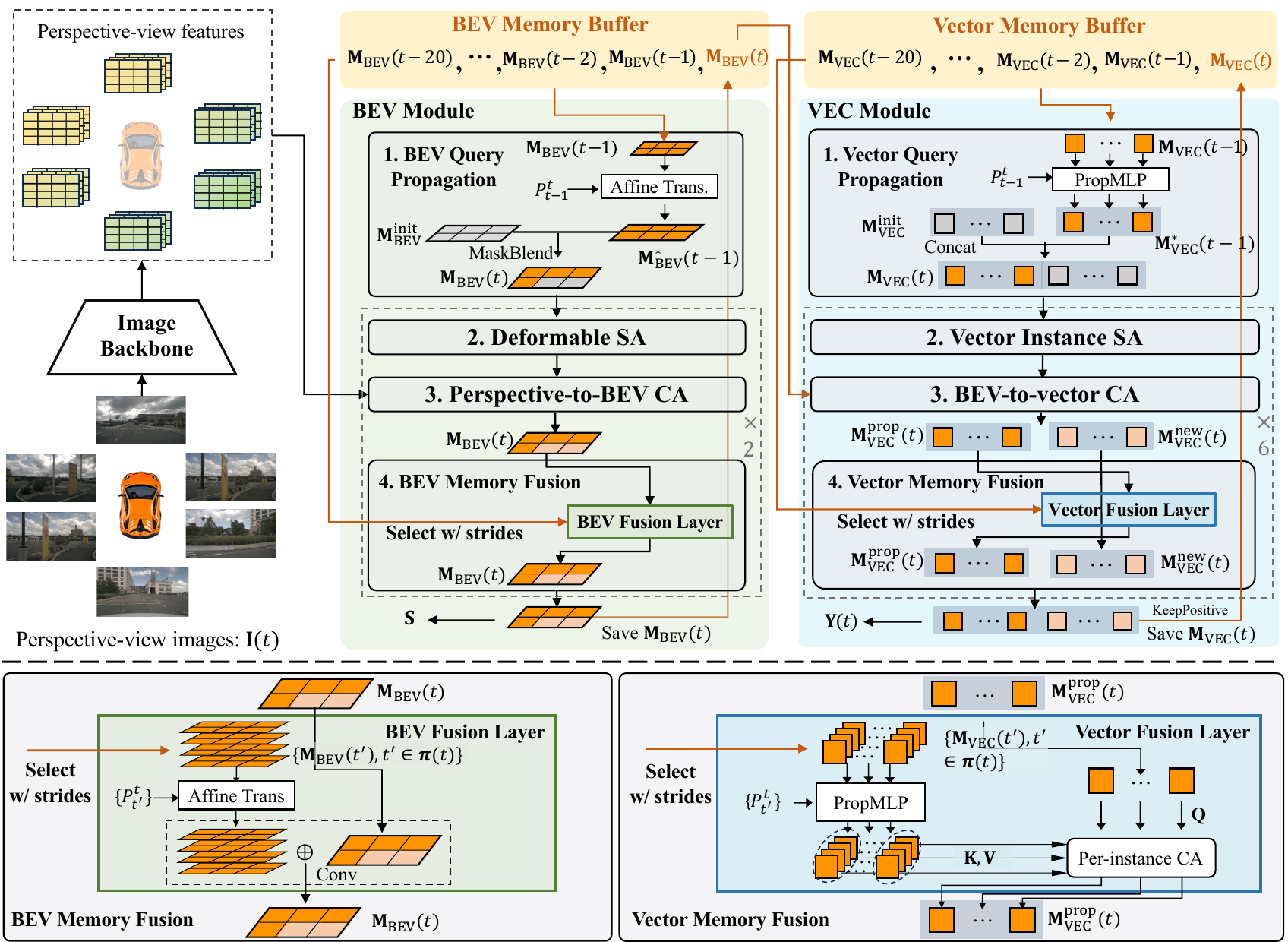}
\caption{\textbf{(Top)} The overall architecture of \ourmethod. \textbf{(Bottom)} The close-up views of the BEV and the Vector fusion layers.}
\label{fig:method_overview}
\end{figure}

%% file: figures/equations.tex
\begin{figure}[!t]
\includegraphics[width=\textwidth]{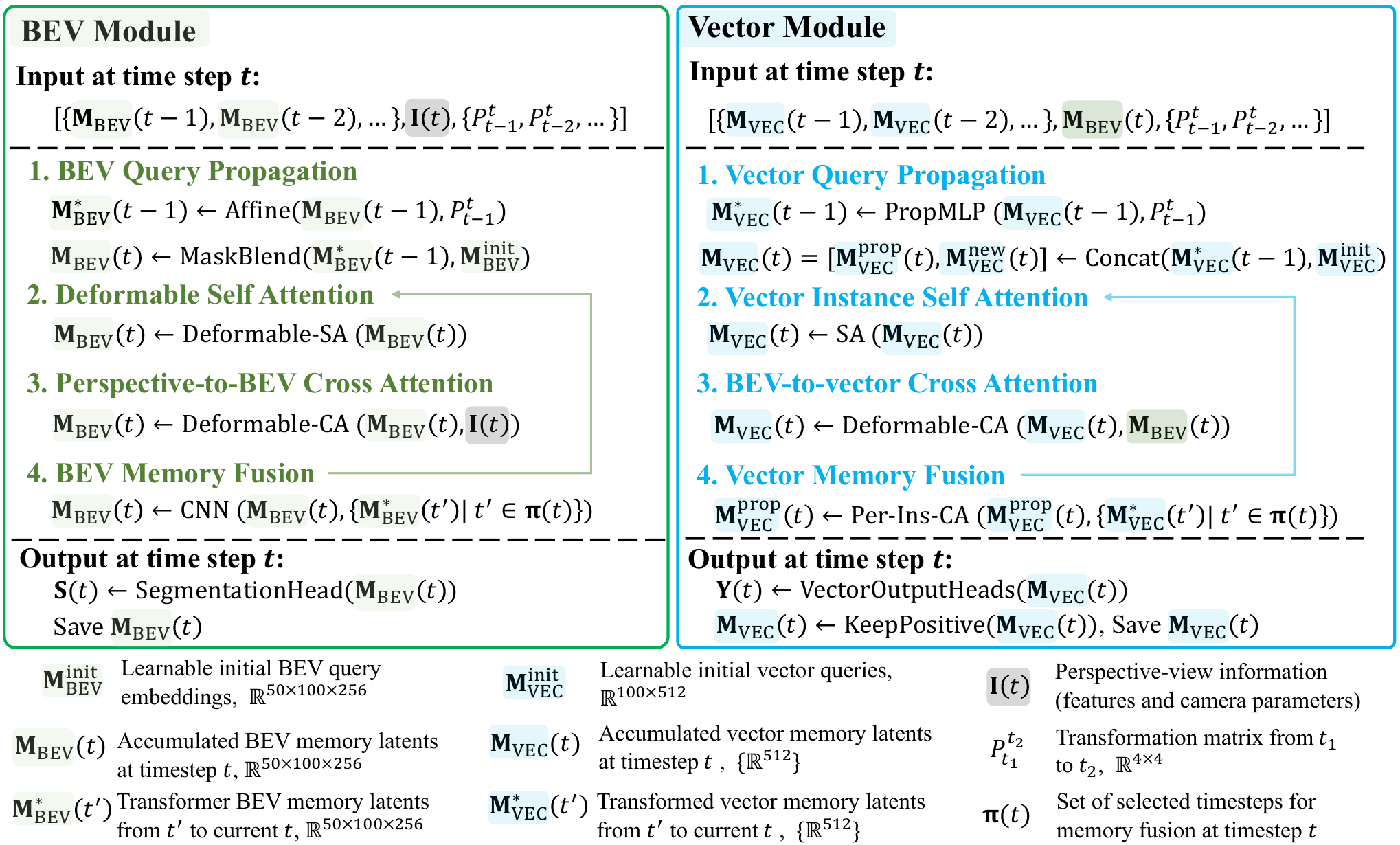}
\caption{The architecture details of the BEV and the Vector modules. 
The BEV-related representations are in \textcolor{ourgreen}{green}, while the vector-related representations are in \textcolor{cyan}{cyan}. Details of the attention layers are described in  \S\ref{sec:method}. 
}
\label{fig:equations}
\end{figure}

%% file: sections/3-evaluation.tex
\section{Consistent Vector HD Mapping Benchmarks}
\label{sec:benchmark}
The section makes existing HD mapping benchmarks consistency-aware by 1) Improving pre-processing to generate temporally consistent ground truth
with ``track'' labels
(\S\ref{subsec:gt_process}); and 2) Augmenting the standard mAP metric with consistency checks (\S\ref{subsec:C-mAP}).

\subsection{Consistent ground truth}
\label{subsec:gt_process}
MapTR~\cite{liao2022maptr,liao2023maptrv2} created vector HD mapping benchmark from nuScenes and Agroverse2 datasets, adopted by many follow-ups~\cite{qiao2023bemapnet,liao2023maptrv2,zhang2023gemap,chen2023polydiffuse,zhang2023mapvr}. However, pedestrian crossings are merged naively and inconsistent across frames. Divider lines are also inconsistent (for Argoverse2) with the failures of its graph tracing process.

StreamMapNet~\cite{yuan2024streammapnet} inherited code from VectorMapNet~\cite{liu2023vectormapnet} and created a benchmark with better ground truth, which has been used in the workshop challenge~\cite{hdmappingWorkshopChallenge}.
However, there are still issues.
For Argoverse2, divider lines are sometimes split into shorter segments. For nuScenes, large pedestrian crossings sometimes split out small loops, whose inconsistencies arise randomly per frame, leading to temporarily inconsistent representations. 
We provide visualizations for the issues of existing benchmarks in the \supp.

We improve processing code from existing benchmarks to (1) enhance per-frame ground-truth geometries, then (2) compute their correspondences across frames, forming ground-truth ``tracks''.

\mypara{(1) Enhancing per-frame geometries}
We inherit and improve the MapTR codebase, which has been popular in the community, while making two changes: Replace the pedestrian-zone processing with the one in StreamMapNet and further improve the quality by more geometric constraints; and Enforce temporal consistency in the divider processing by augmenting the graph tracing algorithm to handle noises of raw annotations (only for Argoverse2).

\mypara{(2) Forming tracks}
Given per-frame road element geometries, we solve an optimal bipartite matching problem between every pair of adjacent frames to establish correspondences of road elements. Pairwise correspondences are chained to form tracks of road elements.
The matching score between a pair of road elements is defined as follows. A road-element geometry is either a polygonal curve or a loop. We transform an element geometry in an older frame to the newer one based on the vehicle motion, then rasterize both curves/loops with a certain thickness into instance masks. Their intersection over union is the matching score. 
Please refer to \supp for the full algorithmic details.

\subsection{Consistency-aware mAP metric}
\label{subsec:C-mAP}
The standard mean average precision (mAP) metric does not penalize temporarily inconsistent reconstructions. We match reconstructed road elements and the ground truth in each frame independently with Chamfer distance, as in the standard mAP process, then remove temporarily inconsistent matches with the following check. First, for baseline methods that do not predict tracking information, we form tracks of reconstructed road elements using the same algorithm we used to get ground-truth temporal correspondences (we also extend the algorithm to re-identify a lost element by trading off the speed; see \supp for details). Next, let an ``ancestor'' be a road element that belongs to the same track in a prior frame. From the beginning of the sequence, we remove a per-frame match (of reconstructed and ground-truth elements) as temporarily inconsistent if any of their ancestors was not a match.
The standard mAP is then calculated with the remaining temporarily consistent matches. See \supp for complete algorithmic details.

%% file: sections/4-exp.tex
\section{Experiments}
\label{sec:exps}
We build our system based on the StreamMapNet codebase, while using 8 NVIDIA RTX A5000 GPUs to train our model for 72 epochs on nuScenes (18, 6, and 48 epochs for the three stages) and 35 epochs on Argoverse2 (12, 3, and 20 epochs for the three stages). The batch sizes for the three training stages are 16, 48, and 16, respectively.
The training takes roughly three days, while the inference speed is roughly 10 FPS.
After explaining the datasets, the metrics, and the baseline methods, the section provides the experimental results.

\mypara{Datasets}
We use the nuScenes~\cite{caesar2020nuscenes} and Argoverse2~\cite{wilson2023argoverse} datasets. nuScenes dataset is annotated with 2Hz with 6 synchronized surrounding cameras. Input perspective images are of size 480$\times$800.
Argoverse2 dataset is annotated with 10 Hz, using 7 surrounding cameras. Input perspective images are of size 608$\times$608.
We follow MapTRv2~\cite{liao2023maptrv2} and use an interval of 4 to subsample the sequences of Argoverse2. We evaluate the methods with the official dataset splits as well as the geographically non-overlapping splits proposed in StreamMapNet~\cite{yuan2024streammapnet}. 

\mypara{Metrics}
We follow prior works~\cite{liu2023vectormapnet,liao2022maptr,liao2023maptrv2,yuan2024streammapnet} and use Average Precision (AP) as the main evaluation metric, where Chamfer distance is the matching criterion. The AP is averaged across three distance thresholds $\{0.5m, 1.0m, 1.5m\}$. The final mean AP (mAP) is computed by averaging the results over the three road element types:
pedestrian crossing, lane-divider, and road-boundary. 
We provide both the original scores and the new consistency-aware augmented scores (\S\ref{subsec:C-mAP}).

\mypara{Baselines}  
MapTRv2~\cite{liao2023maptrv2} and StreamMapNet~\cite{yuan2024streammapnet} are the main baselines due to their popularity and superior performance. 
We run their official codebase and train the models until complete convergence.
The results of recent competing methods~\cite{ding2023pivotnet,zhang2023gemap,wang2024sqd-mapnet} are also included for reference by copying numbers from their corresponding papers.

\input{tables/nuscenes_results}

\subsection{Quantitative evaluations}
\label{subsec:exp:quantitative}
One of our contributions is the temporarily consistent ground truth (GT) over the two existing counterparts (i.e., MapTR~\cite{liao2022maptr,liao2023maptrv2} and StreamMapNet~\cite{yuan2024streammapnet}).
\autoref{tab:nuscenes_main} and \autoref{tab:argo_main} show the results where a system is trained and tested on one of the three GTs  (shown in the first column). Since our codebase is based on StreamMapNet, we evaluate our system on the StreamMapNet GT and our temporarily consistent GT.

\mypara{nuScenes results}
\autoref{tab:nuscenes_main} shows that both MapTRv2 and StreamMapNet achieve better mAP with our GT, which is expected as we fixed the inconsistencies in their original GT (explained in \S\ref{subsec:gt_process}). StreamMapNet's improvement is slightly higher since it has temporal modeling (whereas MapTR does not) and exploits temporal consistency in the data.
\ourmethod significantly outperforms the competing methods, especially with our consistent GT by more than 
8\% and 22\% in the original and the consistency-aware mAP scores.
Note that \ourmethod is the only system to produce explicit tracking information (i.e., correspondences of reconstructed elements across frames), which is required for the consistency-area mAP.
A simple matching algorithm creates tracks for the baseline methods (See \supp for details). We also measure the running speed of 

\input{tables/argo_results}

\mypara{Argoverse2 results}
\autoref{tab:argo_main} shows that both MapTRv2 and StreamMapNet achieve better mAP scores with our consistent GT, which has higher quality GT (for
pedestrian crossings and dividers) besides being temporarily consistent, benefiting all the methods.
\ourmethod outperforms all the other baselines by significant margins (\ie, 11\% or 8\%, respectively) in all settings. The consistency-aware score (C-mAP) further demonstrates our superior consistency, showing an improvement of more than 18\% over StreamMapNet.

\input{tables/newsplit_results}
\label{subsec:exp:newsplit}

\subsection{Results with geographically non-overlapping data}
Official training/testing splits of nuScenes and Agroverse2 datasets have geographical overlaps (i.e., the same roads appear in both training/testing), which allows overfitting~\cite{lilja2023localization-eval-hdmap}.
\autoref{tab:newsplit} compares the best baseline method StreamMapNet~\cite{yuan2024streammapnet} and \ourmethod based on the geographically non-overlapping splits with two different perception ranges (the settings are proposed by StreamMapNet).
\ourmethod consistently outperforms StreamMapNet with significant margins, demonstrating robust cross-scene generalization. Note that
the performance for nuScenes degrades for both methods. Upon careful inspection, road elements were detected successfully, but the regressed coordinates had large errors, leading to low performance.

\subsection{More analysis}
\label{subsec:exp:analysis}

\input{tables/ablations}

\mypara{Ablation study of core model components} 
\autoref{tab:ablations} demonstrate the contributions of key design elements in \ourmethod.
The first ``baseline'' entry is StreamMapNet without its temporal reasoning capabilities (i.e., without its BEV and vector streaming memories and modules). The second entry is StreamMapNet. Both methods are trained for 110 epochs till full convergence.
The last three entries are the variants of \ourmethod with or without the key elements.
The first variant drops the memory fusion components in the BEV/VEC modules. This variant utilizes the tracking formulation but relies on a single BEV/VEC memory to hold the past information, like the GRU embedding of StreamMapNet.
The second variant adds the memory buffers and the memory fusion components but without the striding strategy, that is, using the latest 4 frames for the fusion. This variant 
significantly boosts performance, demonstrating the effectiveness of our memory mechanism.
The last variant adds memory striding, which makes more effective use of the memory mechanism and improves performance.

\input{tables/strides_ablation}
\mypara{Choice of memory strides}
\autoref{tab:ablation_stride} presents the ablation study over the choice of the four distance strides. We believe \{1m, 5m, 10m, 15m\} works well for two main reasons: 1) The first entry (\ie, 1m) helps stabilize the prediction of the current frame and should be close to the current frame to align with the training setup (for each training sample, we randomly sample 4 previous frames from the last 10 frames to form a training clip); and 2) The entries should have proper distance gaps to minimize the information redundancy in the memory.

\input{figures/qualitative}

\mypara{Inference speed}
We evaluate the inference speed of MapTRv2, StreamMapNet, and \ourmethod on the nuScenes dataset using a single NVIDIA RTX 6000 GPU with batch size 1. The results are 12.5 FPS, 14.2 FPS, and 11.5 FPS, respectively. \ourmethod has different query propagation from StreamMapNet and introduces additional memory fusion layers, so the FPS is lower than the original StreamMapNet by 19\% (11.5 vs. 14.2). Our current implementation of the vector memory fusion layer uses for-loops to iterate through all vector instances without batching, which is sub-optimal and could be optimized for better running efficiency.

\subsection{Qualitative evaluations}

\autoref{fig:qualitative} presents qualitative comparisons of \ourmethod and the baseline methods on both nuScenes and Argoverse2 datasets. 
For better visualization, we use a simple algorithm to merge per-frame vector HD maps into a global vector HD map. Please refer to \supp for the details of the merging algorithm and the visualization of per-frame reconstructions.
\ourmethod produces much more accurate and cleaner results, demonstrating superior overall quality and temporal consistency.
For scenarios where the vehicle is turning or not trivially moving forward (including the two examples in \autoref{fig:teaser}), StreamMapNet and MapTRv2 can produce unstable results, thus leading to broken and noisy merged results. This is mainly because the detection-based formulation has difficulties maintaining temporally coherent reconstructions under complex vehicle motions.

%% file: tables/nuscenes_results.tex
\begin{table}[t]
\centering
\small
\setlength{\tabcolsep}{2.5pt}
\caption{Results
on nuScenes\cite{caesar2020nuscenes}. The first column shows three different ground truth used for training and testing. ``Consistent'' is our temporarily consistent ground truth. The standard AP scores are reported for pedestrian crossing, lane-divider, road-boundary, and their average. C-mAP is our consistency-aware metric, which requires tracking information in the ground truth and is reported only for Consistent.
$^+$: Numbers are from the original papers. $^\dagger$: Epochs for our multi-frame training. }
\begin{tabular}{l|lcl|cccc|c}
\toprule
G.T. data & Method & Backbone & Epoch & AP$_{\textit{p}}$ & AP$_{\textit{d}}$ & AP$_{\textit{b}}$ & mAP & C-mAP\\
\midrule
\multirow{4}*{\shortstack[c]{MapTR}} & MapTR$^+$\cite{liao2022maptr} & R50 & 110 & 56.2 & 59.8 & 60.1 & 58.7 & \multirow{4}*{\shortstack[c]{-}}  \\
& PivotNet$^+$~\cite{ding2023pivotnet} & SwinT & 110          
         &  62.6 & 68.0 & 69.7 & 66.8  \\
& MapTRv2$^+$\cite{liao2023maptrv2} & R50  & 110 &  68.1 & 68.3 & 69.7 & 68.7  \\

& GeMap$^+$~\cite{zhang2023gemap} & R50 & 110 & 67.1 & 69.8 & 71.4 & 69.4 \\

\midrule

\multirow{3}*{\shortstack[c]{StmMapNet}} & StreamMapNet\cite{yuan2024streammapnet} & R50 & 110 & 68.0 & 71.2 & 68.0 & 69.1 & \multirow{3}*{\shortstack[c]{-}}   \\
& SQD-MapNet$^+$~\cite{wang2024sqd-mapnet} & R50 & 24 & 63.6 & 66.6 & 64.8 & 65.0 \\
& \ourmethod (Ours) & R50 & 72$^\dagger$ & 77.3 & 72.4 & 74.2 & 74.7 & \\

\midrule

\multirow{3}*{\shortstack[c]{Consistent}} & MapTRv2\cite{liao2023maptrv2} & R50 & 110 & 69.6 & 68.5 & 70.3 &  69.5 & 50.5 \\

& StreamMapNet\cite{yuan2024streammapnet} & R50 & 110 & 70.0 & 72.9 & 68.3 & 70.4 & 56.4 \\
& \ourmethod (Ours) & R50 & 72$^\dagger$ & 80.0 & 74.1 & 74.1 & 76.1 & 69.1 \\
\end{tabular}
\label{tab:nuscenes_main}
\end{table}

%% file: tables/argo_results.tex
\begin{table}[t]
\centering
\small
\setlength{\tabcolsep}{2pt}
\caption{
Results 
on Argoverse2~\cite{wilson2023argoverse}. $^+$: Numbers are from the original papers. $^\dagger$: Epochs for our multi-frame training.
}
\begin{tabular}{l|lcl|cccc|c}
\toprule
G.T. data & Method & Backbone & Epoch & AP$_{\textit{p}}$ & AP$_{\textit{d}}$ & AP$_{\textit{b}}$ & mAP & C-mAP\\
\midrule
\multirow{2}*{\shortstack[c]{MapTR}} & MapTRv2$^+$\cite{liao2023maptrv2} & R50 & 6*4 & 62.9 & 72.1 & 67.1 & 67.4 & \multirow{2}*{\shortstack[c]{-}}   \\
& GeMap$^+$\cite{zhang2023gemap} & R50 & 24*4 & 69.2 & 75.7 & 70.5 & 71.8 &  \\

\midrule

\multirow{4}*{\shortstack[c]{StmMapNet}} & StreamMapNet$^+$\cite{yuan2024streammapnet} & R50 & 30 & 62.0 & 59.5 & 63.0 & 61.5 & \multirow{4}*{\shortstack[c]{-}} \\
& StreamMapNet\cite{yuan2024streammapnet} & R50 & 72 & 65.0 & 62.2 & 64.9 & 64.0 & \\
& SQD-MapNet$^+$\cite{wang2024sqd-mapnet} & R50 & 30 & 64.9 & 60.2 & 64.9 & 63.3 &  \\
& \ourmethod (Ours) & R50 & 35$^\dagger$ & 74.5 & 66.4 & 73.4 & 71.4 \\
\midrule
\multirow{3}*{\shortstack[c]{Consistent}} &
MapTRv2\cite{liao2023maptrv2} & R50 & 24*4 & 68.3 & 75.6 & 68.9 & 70.9 & 56.1   \\
& StreamMapNet\cite{yuan2024streammapnet} & R50 & 72 & 70.5 & 74.2 & 66.1 & 70.3 & 57.5  \\

& \ourmethod (Ours) & R50 & 35$^\dagger$ & 77.0 & 80.0 & 73.7 & 76.9 & 68.3 \\

\end{tabular}
\label{tab:argo_main}
\end{table}

%% file: tables/newsplit_results.tex
\begin{table}[t]
\centering
\small
\setlength{\tabcolsep}{2pt}
\caption{
Results with
geographically non-overlapping data proposed in StreamMapNet~\cite{yuan2024streammapnet}. Our consistent ground truth is used. $^\dagger$: Epochs for our multi-frame training.
}
\resizebox{\textwidth}{!}{
\begin{tabular}{c|c|lcl|ccc|c}
\toprule
Range & Dataset & Method & Epoch & AP$_{\textit{p}}$ & AP$_{\textit{d}}$ & AP$_{\textit{b}}$ & mAP & C-mAP\\
\midrule
\multirow{4}*{\shortstack[c]{60$\times$30m}}& \multirow{2}*{\shortstack[c]{nuScenes\cite{caesar2020nuscenes}}} &
StreamMapNet~\cite{yuan2024streammapnet} & 110 & 31.6 & 28.1 & 40.7 & 33.5 & 22.2\\
& & \ourmethod(Ours) & 72$^\dagger$ & 45.9 & 30.0 & 45.1 & 40.3 & 32.5  \\
\cmidrule{2-9}
& \multirow{2}*{\shortstack[c]{Argoverse2\cite{wilson2023argoverse}}} & StreamMapNet\cite{yuan2024streammapnet}  & 72 & 61.8 & 68.2 & 63.2 & 64.4 & 54.4 \\
& & \ourmethod(Ours) & 35$^\dagger$ & 70.0 & 75.1 & 68.9 & 71.3 & 63.2 \\

\midrule

\multirow{4}*{\shortstack[c]{100$\times$50m}}& \multirow{2}*{\shortstack[c]{nuScenes\cite{caesar2020nuscenes}}} &
StreamMapNet~\cite{yuan2024streammapnet} & 110 & 25.1 & 18.9 & 25.0 & 23.0  & 14.6 \\
& & \ourmethod(Ours) & 72$^\dagger$ & 45.9 & 24.3 & 38.4 & 36.2 & 27.5  \\

\cmidrule{2-9}

& \multirow{2}*{\shortstack[c]{Argoverse2\cite{wilson2023argoverse}}} & StreamMapNet\cite{yuan2024streammapnet}  & 72 & 60.1 & 56.1 & 47.5 & 54.6 & 41.3 \\
& & \ourmethod(Ours) & 35$^\dagger$ & 71.2  & 64.6 & 58.5 & 64.8 & 55.7 \\

\end{tabular}
}
\label{tab:newsplit}
\end{table}

%% file: tables/ablations.tex
\begin{table}[t!]
\centering
\small
\caption{Ablation studies on the key design elements of \ourmethod, evaluated on the nuScenes dataset with our consistent ground truth.}
\resizebox{\linewidth}{!}{
\begin{tabular}{ll ccc ccccc}
\toprule
\setlength{\tabcolsep}{2.5pt}
\multirow{2}{*}{Method} & \multirow{2}{*}{Task} & \multicolumn{3}{c}{Memory}       & \multicolumn{5}{c}{Metrics} \\ 
\cmidrule(r){3-5}\cmidrule(r){6-10}
 &  & Embed. & +Fusion & +Stride &  AP$_{\textit{p}}$ & AP$_{\textit{d}}$ & AP$_{\textit{b}}$ & mAP & C-mAP \\
\midrule
Baseline\cite{yuan2024streammapnet} & Detection & - & - & - & 69.5 & 71.7 & 68.5 & 69.9 & 56.1 \\ 
StmMapNet\cite{yuan2024streammapnet} & Cond. detect. & \cmark & - & - & 70.0 & 72.9 & 68.3 & 70.4 & 56.4 \\
\hdashline
\multirow{3}{*}{\ourmethod} & \multirow{3}{*}{Tracking} & \cmark & - & - & 73.8 & 69.2 & 69.4 & 70.8 & 62.4 \\
& & \cmark & \cmark & - & 78.6 & 73.3 & 72.8  & 74.9 & 68.1 \\
  & & \cmark & \cmark & \cmark & 80.0 & 74.1 & 74.1 & 76.1 & 69.1 \\
\bottomrule
\end{tabular}
}
\label{tab:ablations}
\end{table}

%% file: tables/strides_ablation.tex
\begin{table}[ht!]
\centering
\small
\setlength{\tabcolsep}{3pt}
\caption{
Ablation study of the four distance strides, evaluated on the old data split of the nuScenes dataset with 60m$\times$30m range.
}
\vspace{-1em}
\begin{tabular}{cc|cccc|c}
\toprule
Buffer size & Strides (m) & AP$_{\textit{p}}$ & AP$_{\textit{d}}$ & AP$_{\textit{b}}$ & mAP & C-mAP\\
\midrule
4 & - & 78.6 & 73.3 & 72.8 & 74.9 & 68.1  \\
\midrule
\multirow{5}*{20} & \{0, 0, 0, 0\} & 78.8 & 73.5 & 72.9 & 75.0 & 68.2  \\
& \{1, 2, 3, 4\} & 79.9 & 74.0 & 73.6 & 75.8 & 68.8 \\
& \{1, 3, 5, 7\} & 79.9 & 74.0 & 73.8 & 75.9 & 68.8  \\
& \{1, 5, 10, 15\} & 80.0 & 74.1 & 74.1 & 76.1 & 69.1 \\
& \{5, 10, 15, 20\} & 77.6 & 71.3 & 72.6 & 73.8 & 51.5 \\
\end{tabular}
\label{tab:ablation_stride}
\end{table}

%% file: figures/qualitative.tex
\begin{figure}[!t]
\centering
\includegraphics[width=\textwidth]{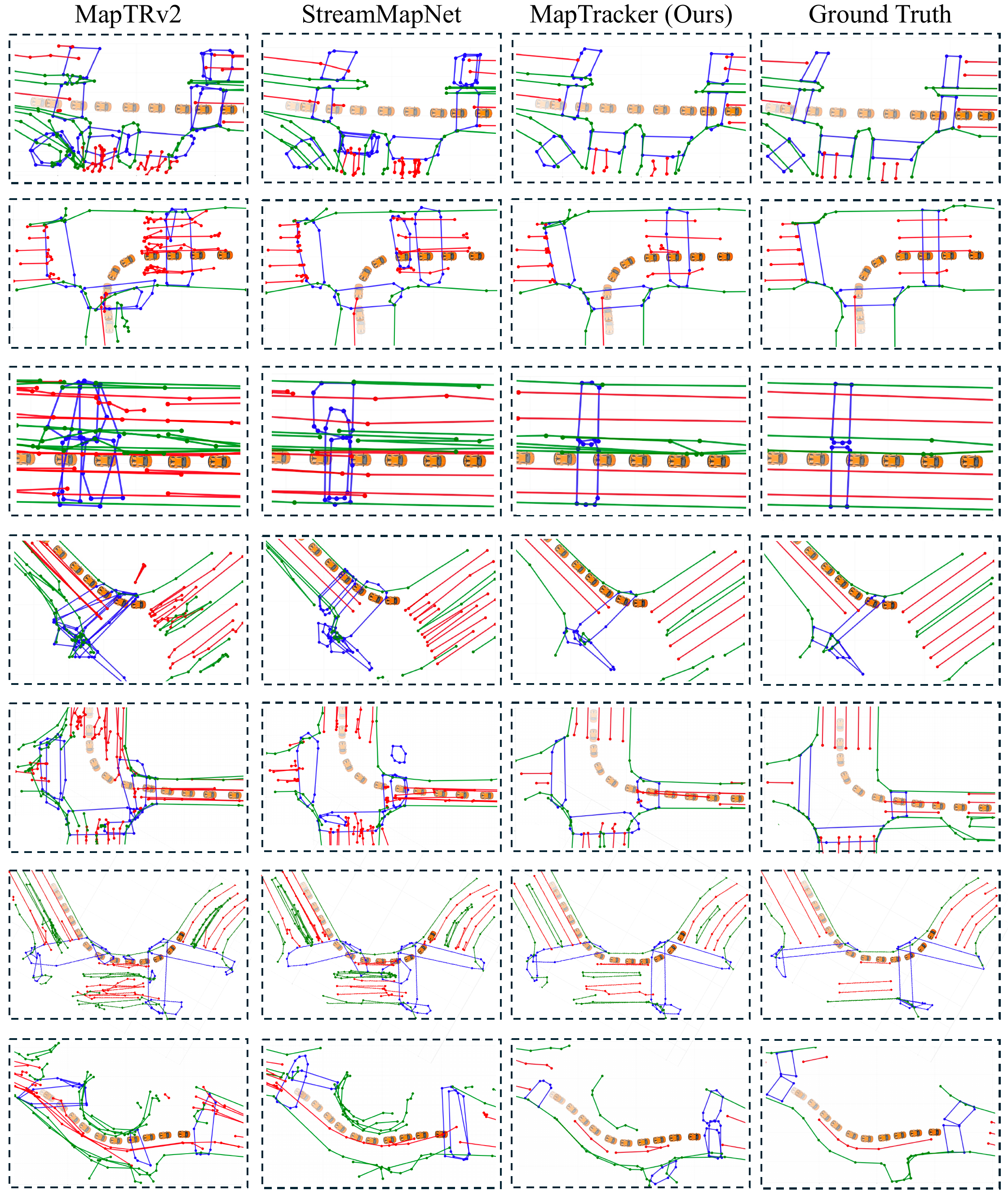}
\vspace{-1.5em}
\caption{Qualitative comparisons of the two representative baselines, \ourmethod(Ours), and the ground truth.
A simple online algorithm merges per-frame vector HD map reconstructions across a single drive-through into a global vector HD map.
The top five examples are from nuScenes, while the bottom two are from Argoverse2. 
}
\label{fig:qualitative}
\end{figure}

%% file: sections/5-conclusion.tex
\section{Conclusion}
This paper introduces MapTracker, which formulates vector HD mapping as a tracking task and leverages a history of raster and vector latents to maintain temporal consistency. We employ a query propagation mechanism to associate tracked road elements across frames, and fuse a subset of memory entries selected with distance strides to enhance consistency.  
We also improve existing benchmarks by generating consistent ground truth with tracking labels and augmenting the original mAP metric with temporal consistency checks.
MapTracker significantly outperforms existing methods on nuScenes and Agroverse2 datasets when evaluated with the traditional metrics and demonstrates superior temporal consistency when evaluated with our consistency-aware metrics.

\mypara{Limitations}
We identify two limitations of \ourmethod. First, the current tracking formulation does not handle the merges and the splits of road elements (\eg, a U-shaped boundary splits into two straight lines in a future frame, or vice versa). The ground truth does not represent them properly either.
Second, our system is still at 10 FPS, falling a bit short of real-time performance, especially at critical crashing events.
Optimizing the efficiency and handling more complex real-world road structures are our future work.

\mypara{Acknowledgements} 
This research is partially supported by NSERC Discovery Grants, NSERC Alliance Grants, and John R. Evans Leaders Fund (JELF). We thank the Digital Research Alliance of Canada and BC DRI Group for providing computational resources.

%% file: sections_supp/supp-implementations.tex
\section{Remaining Details of \ourmethod}
\label{supp:method_details}

This section explains the remaining details of \ourmethod (refer \S3 of the main paper).

\subsection{Transformation loss details}
The transformation loss $\mathcal{L}_\text{trans}$ trains the PropMLP to ensure the latent transformation maintains the geometry and class type. The inputs and outputs of the PropMLP are described in \S3.3, Figure 2, and Figure 3 of the main paper. For the propagated vector latents $\memvec^*(t-1)$ from $t-1$ to $t$, we apply the vector output heads to get the predictions $\vecOutputs^*(t-1)$. We then derive the ground truth $\hat{\vecOutputs}^*(t-1)$ by directly applying the transformation matrix to the ground truth $\hat{\vecOutputs}(t-1)$. Since we have the optimal bipartite match $\matching(t-1)$ of $t-1$, the transformation loss is defined by
\begin{align*}
\mathcal{L}_\text{trans} &= \mathcal{L}_\text{focal}(\{ \hat{c}^*_i(t-1)\}|_{\matching(t-1)}, \{ p_i^*(t-1) \} ) \\
&+ \mathcal{L}_\text{line}(\{ \hat{\curve}_i^*(t-1)\}|_{\matching(t)}, \{ \curve_i^*(t-1) \} ),
\end{align*}
where $\hat{\vecOutputs}^*(t-1) = \{ \hat{\curve}_i^*(t-1), \hat{c}^*_i(t-1) \}$ and $\vecOutputs^*(t-1) = \{ \curve_i^*(t-1), p^*_i(t-1) \}$.

\subsection{Memory fusion details} 

\S3.2 and \S3.3 of the main paper have explained our BEV and vector memory fusion. We provide more implementation details here.

\mypara{Strided memory selection} We present the concrete implementation steps of our strided memory selection in the following. Firstly, we sort all memory entries based on the distance to the current location. Then, we select one closest memory entry for each stride value, starting from the farthest stride (\ie, 15m). If the memory buffer contains less than four history latents, we simply take all.

\mypara{Vector fusion layer} For the per-instance cross-attention of the vector fusion layer, we compute the absolute value of the relative frame difference from the history to the current frame, encode it with the sin/cos positional encoding, and use the encoding as the position encoding for the key and values of the cross-attention.

\section{Remaining Details of Benchmark Contributions}
\label{supp:consistent_benchmark}

This section presents thorough details of our consistent vector HD mapping benchmarks, complementing \S4 of the main paper.

\input{figures/fig_supp_ped_gt_fix}

\subsection{Consistent ground truth}
\label{supp:consistent_gt}

We review the typical problems of the two existing ground-truth data and demonstrate the improved quality of our consistent ground truth.

\input{figures/fig_supp_divider_gt_fix}

\mypara{Pedestrain-crossing (nuScenes and Argoverse2)} \autoref{fig:ped_failure} shows typical failure cases in MapTR~\cite{liao2023maptrv2} and StreamMapNet's~\cite{yuan2024streammapnet} ground truth (GT) for the pedestrian crossing class. The examples are from the nuScenes dataset, where the raw annotations are many small polygon pieces. A merging algorithm must merge the pieces to get a correct global polygon for each pedestrian crossing instance.  MapTR's merging algorithm merges all small polygons with overlaps in a brute-force way, making some polygons merge and split when crossing the perception boundary and introducing temporal inconsistency. StreamMapNet's merging algorithm considers the orientation of each polygon piece to avoid merging orthogonal pedestrian crossings, thus achieving better temporal consistency. However, the algorithm sometimes fails to handle noisy small pieces near the perception boundary. We borrow the merging algorithm from StreamMapNet and impose more geometric constraints as conditions when merging the polygon pieces, resulting in almost perfect temporal consistency. 

In Argoverse2, the raw annotations of each pedestrian crossing are two line segments, making the ground truth processing easier. However, MapTR and StreamMapNet's processing codes sometimes produce open-loop curves at the perception boundary. We fix their codes to always produce closed polygon loops.

\mypara{Divider (Argoverse2)}
\autoref{fig:divider_failure} shows the failure cases of the lane divider class in MapTR and StreamMapNet's ground truth. The examples are from Argoverse2, where the raw annotations are many short divider segments, and a merging algorithm should merge the segments belonging to the same divider instance.
MapTR employs a graph-tracing algorithm to connect the line segments, where each segment is a node, and segments of the same instance are connected by tracing from a root to the leaf. However, some annotations are corrupted with incomplete graph information, making the graph tracing algorithm fail completely and miss entire dividers. 
For StreamMapNet, owing to its unstable threshold-based rules to connect divider segments, it sometimes fails to produce correct long dividers, leading to temporal inconsistency. 

\input{algorithms/track_generation}

To obtain better ground truth, we first fix the graph-tracing algorithm to avoid missing entire dividers. To handle the noisy/corrupted graph information from the annotations, we then borrow the threshold-based rules from StreamMapNet to further connect the dividers produced by the graph tracing algorithm.

\subsection{Track extraction algorithm}
\label{supp:track_extract}
We show details of the track extraction algorithm in \autoref{alg:track-generation}. This algorithm forms tracks for 1) our consistent ground truth to generate the temporal alignments and 2) the baseline methods to form tracks from per-frame reconstructions. The algorithm has a ``look-back'' hyper-parameter $N$, specifying how many previous frames to check when determining the temporal correspondence (\ie, assigning a global ID to an element in the current frame). Larger look-back parameters better tolerate missing reconstructions by re-identification but greatly slow down the entire vector HD mapping pipeline.

The ground-truth track formation uses $N=1$ (See \S4.1 of the main paper).
In the main paper, the baselines use $ N=1$ to have similar real-time inference speeds for fair evaluations.
A large $N$ improves the C-mAP metric for all the methods but is computationally expensive due to the per-instance rasterization and the bipartite matching. For example, the track extraction algorithm alone is almost 4 times more expensive than the MapTRv2 baseline (with ResNet50) when $N=5$. As reference, this appendix (\S\ref{supp:subsec:full-cmap}) provides additional results when $N=3$ or $N=5$.

\input{algorithms/consist_map}

\subsection{Consistent-aware mAP}
\label{supp:c-map}

\autoref{alg:c-map} presents the algorithmic details for computing our Consistent-aware mAP (C-mAP). Note that the algorithm computes the C-mAP of one distance threshold, while the actual evaluation metrics compute the results of three distances and take the average. The consistency check is at the line 11 of the algorithm.
The definition of the C-mAP metric does not include the track extraction algorithm and does not introduce extra hyperparameters. An ideal vector HD mapping method should explicitly predict tracks of reconstruction like \ourmethod, instead of relying on an external algorithm for track formation.

The conventional distance-based mAP (reported in the main paper) considers all predictions, including those with low confidence scores, when computing the area under curve to get per-class average precision (AP). However, since the tracks are only defined for positive predictions (negative predictions do not have a ``global ID''), the computation of C-mAP excludes negative predictions. Therefore, the value of C-mAP can never reach the conventional mAP, even with perfectly predicted temporal correspondences.
To set up an upper bound for C-mAP when all consistency checks are passed, \S\ref{supp:subsec:full-cmap} reports the results of ``$\overline{\text{C-mAP}}$'', which computes C-mAP by ignoring the consistency checks.

\input{algorithms/merge}

\section{Details of Online Merging Algorithm}
\label{supp:online_merging}

\autoref{alg:merging} presents the high-level pseudo-code for our online merging algorithm, which merges per-frame reconstructions into a global vector HD map. Note that our merging algorithm is simple and not perfect, and may occasionally fail to accurately merge the per-frame data. %
However, the main goal of implementing this merging algorithm is to investigate and analyze the consistency of per-frame reconstructions. More advanced algorithms can be designed and implemented if we need high-quality merged global maps.

%% file: figures/fig_supp_ped_gt_fix.tex
\begin{figure}[t]
\centering
\includegraphics[width=\textwidth]{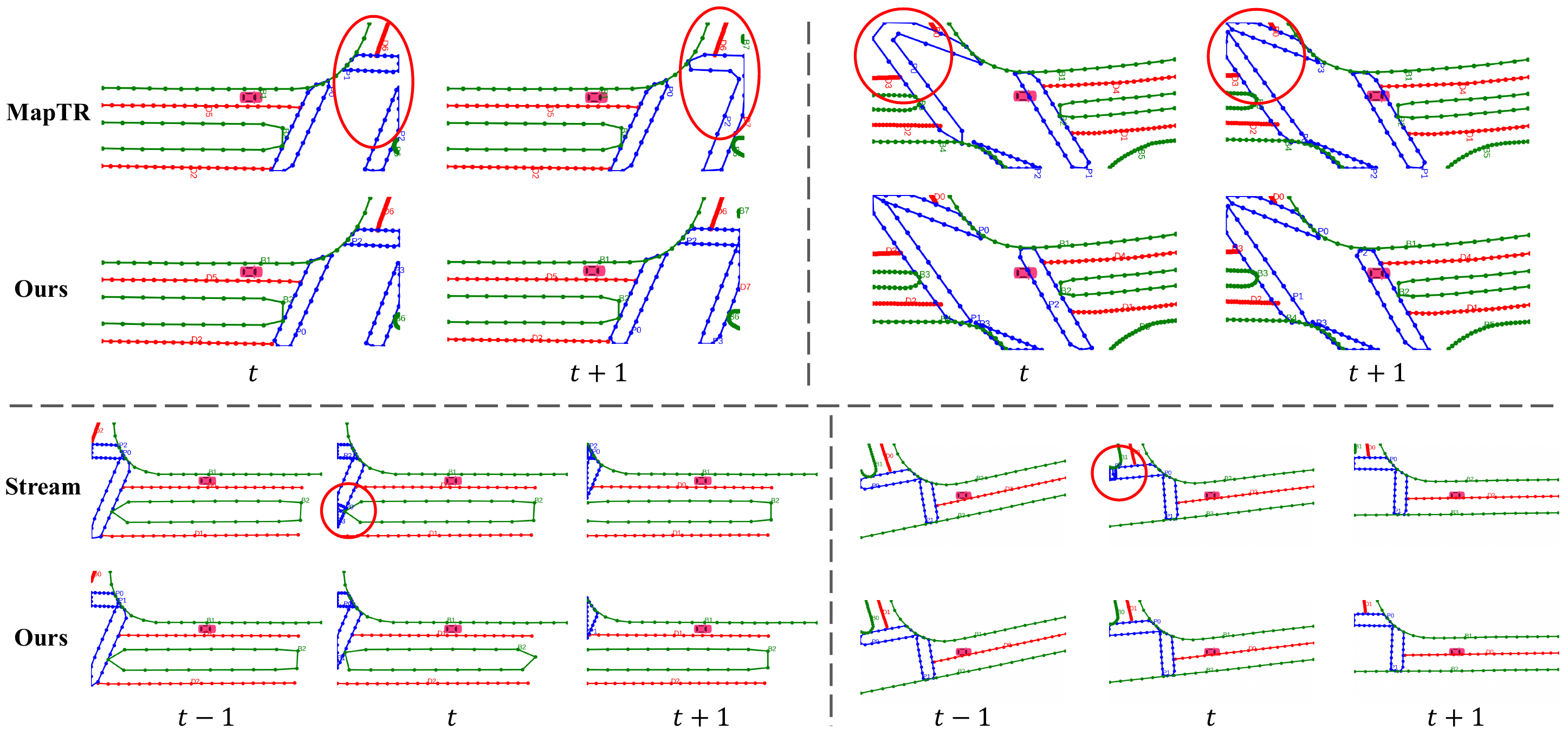}
\caption{
Typical examples of problematic pedestrian crossing annotations in existing ground truth for nuScenes. (Top) MapTR's ground truth merges or splits nearby pedestrian crossings at the perception boundary, leading to temporal inconsistencies. (Bottom) StreamMapNet's ground truth does not have the above merge/split issue but sometimes fails to fuse small polygons (from raw annotations) into a global one. 
}
\label{fig:ped_failure}
\end{figure}

%% file: figures/fig_supp_divider_gt_fix.tex
\begin{figure}[!t]
\centering
\includegraphics[width=\textwidth]{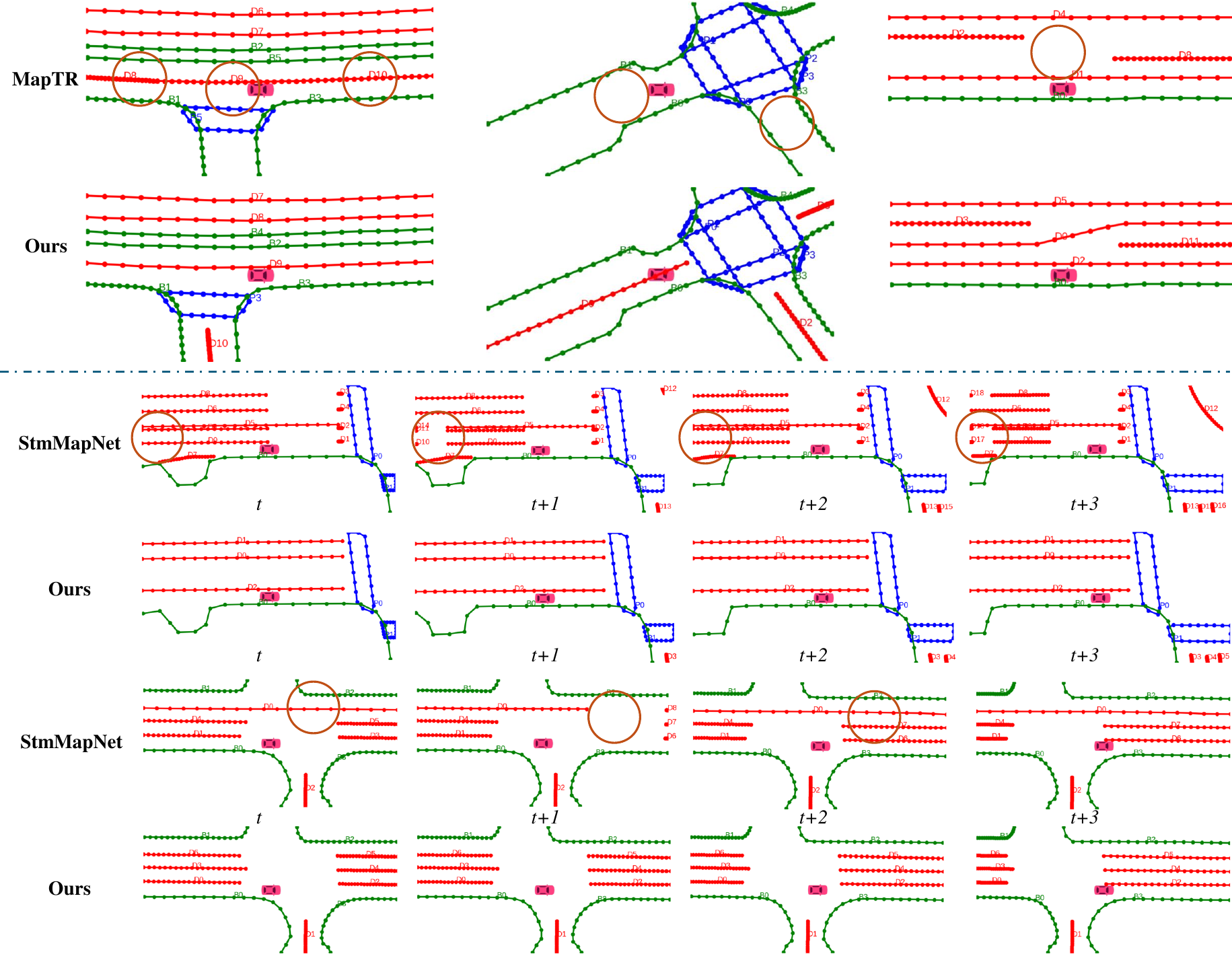}
\caption{Typical failure cases in MapTR and StreamMapNet's ground truth dividers on Argoverse2. (Top) MapTR's ground truth fails to properly merge short divider segments into a global one, and the entire dividers can even be missing due to the failure of its graph tracing algorithm. (Bottom) StreamMapNet's ground-truth dividers suffer from temporal inconsistency (split and merge constantly as the car goes).}
\label{fig:divider_failure}
\end{figure}

%% file: algorithms/track_generation.tex
\begin{algorithm}[!t]
\caption{Track Generation Algorithm}
\label{alg:track-generation}
\begin{algorithmic}[1]
    \STATE \textbf{Input:} Sequence of predicted vectors $\{V(t), t=1,...,T \}$,  Sequence of predicted scores $\{p(t), t=1,...,T \}$, filter threshold $\tau=0.4$, look-back frame number $N$

    \STATE \textbf{Output:}
    $\{M_\text{ID}(t), t=1,...T\}$,  $M_\text{ID}(t)$ records the global id of positive predictions in $V(t)$
    
    \FOR{$t = 1 : T$}
        \STATE Init $M_\text{ID}(t)$ as an empty mapping
        \STATE Obtain a subset of positive vectors $V'(t) = \{ V_j(t) \in V(t), p_j(t) > \tau \}$
        \STATE \textbf{if} $t=0$ \textbf{then}
        \STATE $~~~~$ Assign a new global id to each $v_j'(t)$, update $M_\text{ID}(t)$
        \STATE $~~~~$ \textbf{continue}
        \STATE \textbf{end if}

        \STATE \textbf{for} $k=1 : N$ \textbf{do}
            \STATE $~~~~$ Transform $V'(t-k)$ to the current frame, $V'_t(t-k) $= Affine($V'(t-k),P_{t-k}^t$) 
            \STATE $~~~~$ Do bipartite matching between $V'(t)$ with $V'_{t}(t-k)$ using the IoU between rasterized masks, store the optimal bipartite matching as $B_{rec}(k)$
        \STATE \textbf{end for}

        \STATE \textbf{for} $k=1: N$ \textbf{do}
            \STATE $~~~~$ \textbf{for} $v'_j(t)$ in $V'(t)$
                \STATE $~~~~$$~~~~$ \textbf{if} $v'_j(t)$ doesn't have a global id in $M_\text{ID}(t)$, and $v'_j(t)$ in $B_{rec}(k)$ \textbf{then}
                    \STATE $~~~~$$~~~~$$~~~~$ Get the global id of $v'_j(t)$'s matched instance in $M_\text{ID}(t-k)$,
                    \STATE $~~~~$$~~~~$$~~~~$ Assign this id to $v'_j(t)$, update $M_\text{ID}(t)$
                \STATE $~~~~$$~~~~$ \textbf{end if}
            \STATE $~~~~$ \textbf{end for}
        \STATE \textbf{end for}

        \STATE \textbf{for} $v'_j(t)$ in $V'(t)$
            \STATE $~~~~$ \textbf{if} $v'_j(t)$ doesn't have a global id in $M_\text{ID}(t)$ \textbf{then}
            \STATE $~~~~$ $~~~~$ Assign a new global id to $v'_j(t)$, update $M_\text{ID}(t)$
            \STATE $~~~~$ \textbf{end if}
        \STATE \textbf{end for}     
    \ENDFOR
\end{algorithmic}
\end{algorithm}

%% file: algorithms/consist_map.tex
\begin{algorithm}[!t]
\caption{Consist-aware mAP}
\label{alg:c-map}
\begin{algorithmic}[1]
    \STATE \textbf{Input:} Predicted vectors $V$, GT vectors $\hat{V}$, Predicted Global ID $I$, GT Global ID $\hat{I}$, Predicted scores $P$, a Chamfer distance threshold $\sigma$ (\eg, 0.5m)
    \STATE \textbf{Output:} The C-mAP on the test set
    
    \FOR{each sequence in the test set}
    \STATE $B_{rec}$ records the matching between predictions and GT across the sequence 
    \FOR{each timestep $t$}
        \STATE Obtain the optimal bipartite matching between $V(t)$ and $\hat{V}(t)$, denoted as $I(t) \rightarrow \hat{I}(t)$, and sort $V(t)$ in descending order based on $P(t)$
        \STATE \textbf{for} ${v}_j(t)$ in $V(t)$
        
            \STATE $~~~~$ \textbf{if} $v_j(t)$ has a matched GT vector with Chamfer distance $\le \sigma$  \textbf{then}
                \STATE $~~~~$$~~~~$ Define $\hat{I}_{j}$ as the global ID of the GT vector that matches $v_j(t)$
                \STATE  $~~~~$$~~~~$ \textbf{if} $\hat{I}_j$ has existed in $B_{rec}$ \textbf{then}
                \STATE $~~~~$$~~~~$$~~~~$ \textbf{if} its matched prediction ID is not $I_j$ // Consistency check
                    \STATE $~~~~$$~~~~$$~~~~$$~~~~$ Consider $v_j(t)$ as {FP} (False Positive)
                \STATE $~~~~$$~~~~$$~~~~$ \textbf{else}
                    \STATE $~~~~$$~~~~$$~~~~$$~~~~$ Consider $v_j(t)$ as {TP} (True Positive) 
                    \STATE $~~~~$$~~~~$$~~~~$ \textbf{end if}
                \STATE $~~~~$$~~~~$ \textbf{else}
                    \STATE $~~~~$$~~~~$$~~~~$ Consider $v_j(t)$ as {TP} (True Positive), and update $B_{rec}$
                
                \STATE $~~~~$$~~~~$ \textbf{end if}
            \STATE $~~~~$ \textbf{else}
                \STATE $~~~~$$~~~~$ Consider $v_j(t)$ as {FP}
            \STATE $~~~~$ \textbf{end if}
            \STATE $~~~~$ Record the TP/FP for $v_j(t)$, along with its score $p_j(t)$
        \STATE \textbf{end for}
        \STATE Get TP, FP, and scores for $V(t)$
    \ENDFOR
    \STATE Get TP, FP, and scores for the entire sequence
    \ENDFOR
    \STATE Sort TP and FP of all sequences with the scores in descending order to calculate the AP, get the consistency-aware AP (C-AP). 
    \STATE The C-mAP is the average C-AP across all classes
\end{algorithmic}
\end{algorithm}

%% file: algorithms/merge.tex
\begin{algorithm}[!t]
    \caption{Online Merging Algorithm}
    \label{alg:merging}
    \begin{algorithmic}[1]
        \STATE \textbf{Input:} Predicted set $Y(t)$ for each timestep $t$;
        \STATE Dictionary $D[I]$ records the merged vectors, $I$ represents the global ID of the predicted vectors
        \FOR{each timestep $t$}
            \FOR{each pair $(V_i, c_i)$ in $Y(t)$}
                \STATE $I_i \leftarrow$ Global ID of $V_i$
                \IF{$I_i$ is not in $D$}
                    \STATE $D[I_i] = V_i$
                    \STATE \textbf{continue}
                \ELSE
                    \IF{$c_i$ == Pedestrian Crossing}
                        \STATE $D[I_i] = \texttt{MergeCrossing}(D[I_i], V_i)$  // Merge crossing by finding the convex hull that contains all points in $D[I_i]$ and $V_i$
                        
                    \ELSIF{$c_i$ == Lane Divider}
                        \STATE $D[I_i] = \texttt{MergeDivider}(D[I_i], V_i)$ // Merge divider by interpolating $D[I_i]$ and $V_i$
                    \ELSIF{$c_i$ == Road Boundary}
                        \STATE $D[I_i] = \texttt{MergeBoundary}(D[I_i], V_i)$ // Merge boundary by interpolating $D[I_i]$ and $V_i$
                    \ENDIF
                \ENDIF
            \ENDFOR
        \ENDFOR
    \end{algorithmic}
\end{algorithm}

%% file: sections_supp/supp-exp.tex
\section{Additional Experimental Results and Analyses}
\label{supp:exp}

This section presents additional experimental results and analyses, complementing \S5 of the main paper.

\input{tables/supp_full_with_track_gen_nusc}

\subsection{Full C-mAP results}
\label{supp:subsec:full-cmap}

As discussed in \S\ref{supp:track_extract}, the track extraction algorithm is more robust to temporal inconsistency in the reconstructions when using higher look-back parameters. \autoref{tab:full_cmap_nusc} and \autoref{tab:full_cmap_av2} in this appendix extend the Table 1 and Table 2 of the main paper by providing C-mAP results with different look-back parameters. The first row of each method is the same as the results reported in the main paper, and \ourmethod directly uses the predicted track. \S\ref{supp:qualitative} contains qualitative results with different look-back parameters. We analyze the results of the three methods below.

\myparapara{MapTRv2} gets huge boosts on C-mAP with increased look-back parameters, especially on the nuScenes dataset. This suggests that MapTRv2 suffers from poor temporal consistency, and the predicted road elements frequently disappear and reappear within 2 or 3 consecutive frames.

\input{tables/supp_full_with_track_gen_av2}

\myparapara{StreamMapNet} also benefits from higher look-back parameters. Note that we tried to derive the tracks from StreamMapNet's hidden query propagation but found almost no temporal correspondence -- very few propagated elements stay positive in the next frame. This is mainly because the detection-based formulation cannot exploit tracking labels, and the model treats the propagated information as extra conditions without capturing explicit temporal relationships. StreamMapNet's C-mAP results are worse than MapTRv2 on Argoverse2, further indicating the limitations of its temporal modeling designs.

\myparapara{\ourmethod} predicts tracks and obtains good results without the track extraction algorithm. Note that the track extraction algorithm and our VEC module use different thresholds for determining positive road elements. Our VEC module uses higher thresholds and output fewer positive elements, leading to slightly lower C-mAP compared to the result of using the track extraction algorithm with $N=1$. 
When increasing the look-back parameters, the C-mAP of \ourmethod can also keep improving -- Although \ourmethod obtains much more consistent reconstructions than the baselines, the predicted temporal correspondences are sometimes incorrect, and road elements are still occasionally unstable (\ie, disappear and reappear within several frames). 

As explained in \S\ref{supp:c-map}, $\overline{\text{C-mAP}}$ is the upper bound of C-mAP when the reconstructions pass all consistency checks. \ourmethod gets very close to the $\overline{\text{C-mAP}}$ when we use the track extraction algorithm with $N=5$ to trade running time for track quality -- The gap (1.8) is much smaller than the gap of the baselines, again demonstrating our superior consistency.

\subsection{Check MapTR ground truth}
\autoref{tab:eval-maptr-data} investigates the quality of MapTR's ground truth by evaluating its data on our consistent benchmarks. The tracks of MapTR's ground truth are extracted using \autoref{alg:track-generation} with look-back parameter 1. The results in the table are consistent with what we have analyzed in \S\ref{supp:consistent_gt}: 1) the pedestrian crossings of MapTR data suffer from temporal inconsistencies on both datasets; 2) the dividers have severe issues (\eg, entirely dropped) on Argoverse2. 

\input{tables/supp_maptr_gt}

According to the table, we observe failure in Pedestrian Crossing for both datasets and severe misalignment in Argoverse2, which matches the conclusion of our observation mentioned in the main paper.

\input{figures/additional_qualitative}

\subsection{More qualitative results}
\label{supp:qualitative}

\autoref{supp:qualitative:1} to \autoref{supp:qualitative:6} present additional qualitative comparisons. We show two additional rows for each example compared to Figure 4 of the main paper: \textbf{(the second row, ``Merged LB5''} is the merged results for all the methods when using the track extraction algorithm with look-back parameter $N=5$; \textbf{(the third row, ``Unmerged'')} is the raw unmerged results that simply place the reconstructions of all frames together. For the first row, \ourmethod uses the predicted tracks while the baselines use the track extraction algorithm with look-back parameter $N=1$. 

Note that since our VEC module and the track extraction algorithm use different thresholds for determining the positive predictions, \ourmethod's results in the first and second rows of each example can be slightly different. The second row of \ourmethod sometimes has more positive road elements than the first row, and we use the predicted tracks to plot the unmerged results.    
Comparing each row of each example, \ourmethod's results are cleaner and more consistent, further validating our contributions toward consistent vector HD mapping.

%% file: tables/supp_full_with_track_gen_nusc.tex
\begin{table}[t]
\centering
\small
\setlength{\tabcolsep}{2pt}
\caption{
Full C-mAP on nuScenes\cite{caesar2020nuscenes}.$^\dagger$: Epochs for our multi-frame training.
}
\begin{tabular}{c cc cccc c}
\toprule
Method &  Epoch &  Lookback & C-AP$_{\textit{p}}$ & C-AP$_{\textit{d}}$ & C-AP$_{\textit{b}}$ & C-mAP & $\overline{\text{C-mAP}}$ \\
\midrule
\multirow{3}*{\shortstack[c]{MapTRv2}} &  \multirow{3}*{\shortstack[c]{110}} & 1 frame & 55.8 & 43.8 & 57.9 &  50.5 &  \multirow{3}*{\shortstack[c]{64.9}} \\
&    &  3 frames & 61.5 & 54.0 & 61.3 &  58.9 \\
&    &  5 frames  &62.5 & 55.4 & 62.2 &  60.0 \\

\midrule
\multirow{3}*{\shortstack[c]{StreamMapNet}} &  \multirow{3}*{\shortstack[c]{110}} & 1 frame & 58.6 & 53.5 & 57.1 & 56.4  &  \multirow{3}*{\shortstack[c]{65.9}}\\
&     & 3 frames & 62.8 & 59.7 & 58.9 & 60.5  \\

&     & 5 frames & 63.1 & 60.8 & 59.3 & 61.0  \\

\midrule
\multirow{4}*{\shortstack[c]{\ourmethod}} &  \multirow{4}*{\shortstack[c]{72$^\dagger$}} & $\varnothing$ (predicted)  &  75.4 & 65.0 & 66.9 &  69.1 & \multirow{4}*{\shortstack[c]{72.5}} \\

&     &  1 frame & 75.5 & 65.9 & 67.6 & 69.7  \\ 
&    &  3 frames & 76.3 & 66.8 & 68.2 & 70.4  \\ 
&    &   5 frames & 76.9 & 67.0 & 68.4 & 70.7  \\ 

\end{tabular}
\label{tab:full_cmap_nusc}
\end{table}

%% file: tables/supp_full_with_track_gen_av2.tex
\begin{table}[t]
\centering
\small
\setlength{\tabcolsep}{2pt}
\caption{
Full C-mAP on Argoverse2\cite{wilson2023argoverse}. $^\dagger$: Epochs for our multi-frame training.
}

\begin{tabular}{c cc cccc c}
\toprule
Method &  Epoch & Lookback & C-AP$_{\textit{p}}$ & C-AP$_{\textit{d}}$ & C-AP$_{\textit{b}}$ & C-mAP & $\overline{\text{C-mAP}}$ \\
\midrule
\multirow{3}*{\shortstack[c]{MapTRv2}} &  \multirow{3}*{\shortstack[c]{24*4}}  & 1 frame & 58.4 & 52.5 & 57.4 & 56.1 &  \multirow{3}*{\shortstack[c]{67.7}} \\
&     & 3 frames & 63.2 & 62.1 & 62.8 &  62.7 \\
&     & 5 frames  & 63.9 & 64.1 & 63.0 &  63.7 \\

\midrule
\multirow{3}*{\shortstack[c]{StreamMapNet}} &  \multirow{3}*{\shortstack[c]{72}} & 1 frame &  63.0 & 53.3 & 56.2 & 57.5 &  \multirow{3}*{\shortstack[c]{65.8}}\\
&     & 3 frames & 65.5 & 58.8 & 58.9 & 61.0  \\
&    &  5 frames & 65.6 & 59.5 & 59.0 & 61.4  \\

\midrule
\multirow{4}*{\shortstack[c]{\ourmethod}} &   \multirow{4}*{\shortstack[c]{35$^\dagger$}} & $\varnothing$ (predicted)  & 70.8 & 68.3 & 66.0 &  68.3 & \multirow{4}*{\shortstack[c]{72.8}} \\

&    &  1 frame & 72.6 & 69.5 & 67.4 & 69.8  \\ 
&     & 3 frames & 73.2 & 71.0 & 68.3 & 70.8  \\ 
&     &  5 frames &73.4 & 71.3 & 68.3 & 71.0  \\ 

\end{tabular}
\label{tab:full_cmap_av2}
\end{table}

%% file: tables/supp_maptr_gt.tex
\begin{table}[!th]
\centering
\setlength{\tabcolsep}{3pt}
\caption{Evaluating MapTR's ground-truth data with our consistent benchmarks. The goal is to understand the temporal consistency of MapTR's ground truth.}
\begin{tabular}{c c c c c}
\toprule
 & C-AP$_{\textit{p}}$ & C-AP$_{\textit{d}}$ & C-AP$_{\textit{b}}$ & C-mAP \\
\midrule
nuScenes    & 80.6 & 100 & 100  &  93.5 \\
Argoverse2    & 95.6 & 65.7 & 100 &  87.1  \\
\bottomrule
\end{tabular}
\label{tab:eval-maptr-data}
\end{table}

%% file: figures/additional_qualitative.tex
\begin{figure}[t]
\centering
\includegraphics[width=\textwidth]{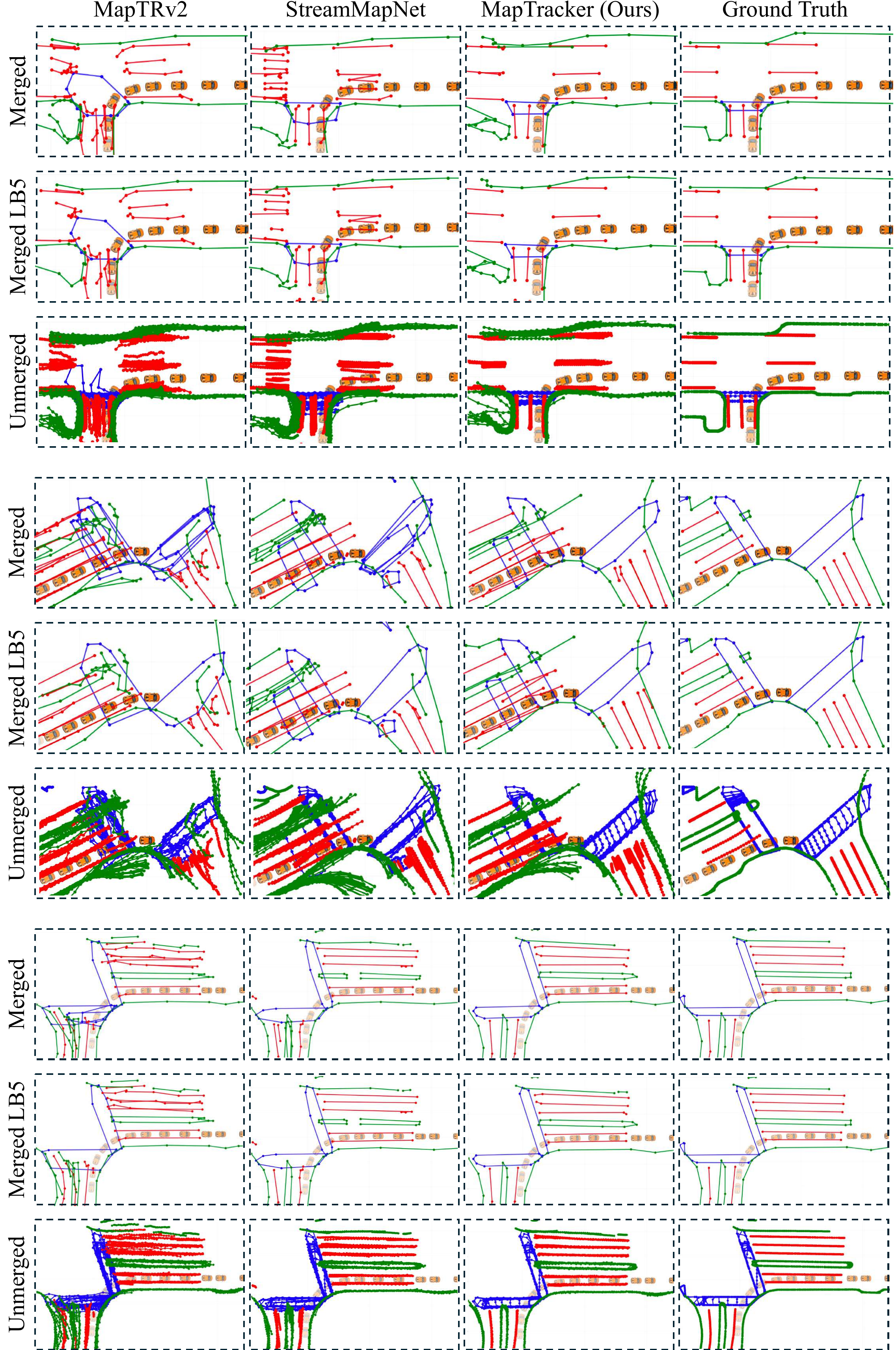}
\caption{Additional qualitative results on the nuScenes dataset.
}
\label{supp:qualitative:1}
\end{figure}

\begin{figure}[t]
\centering
\includegraphics[width=\textwidth]{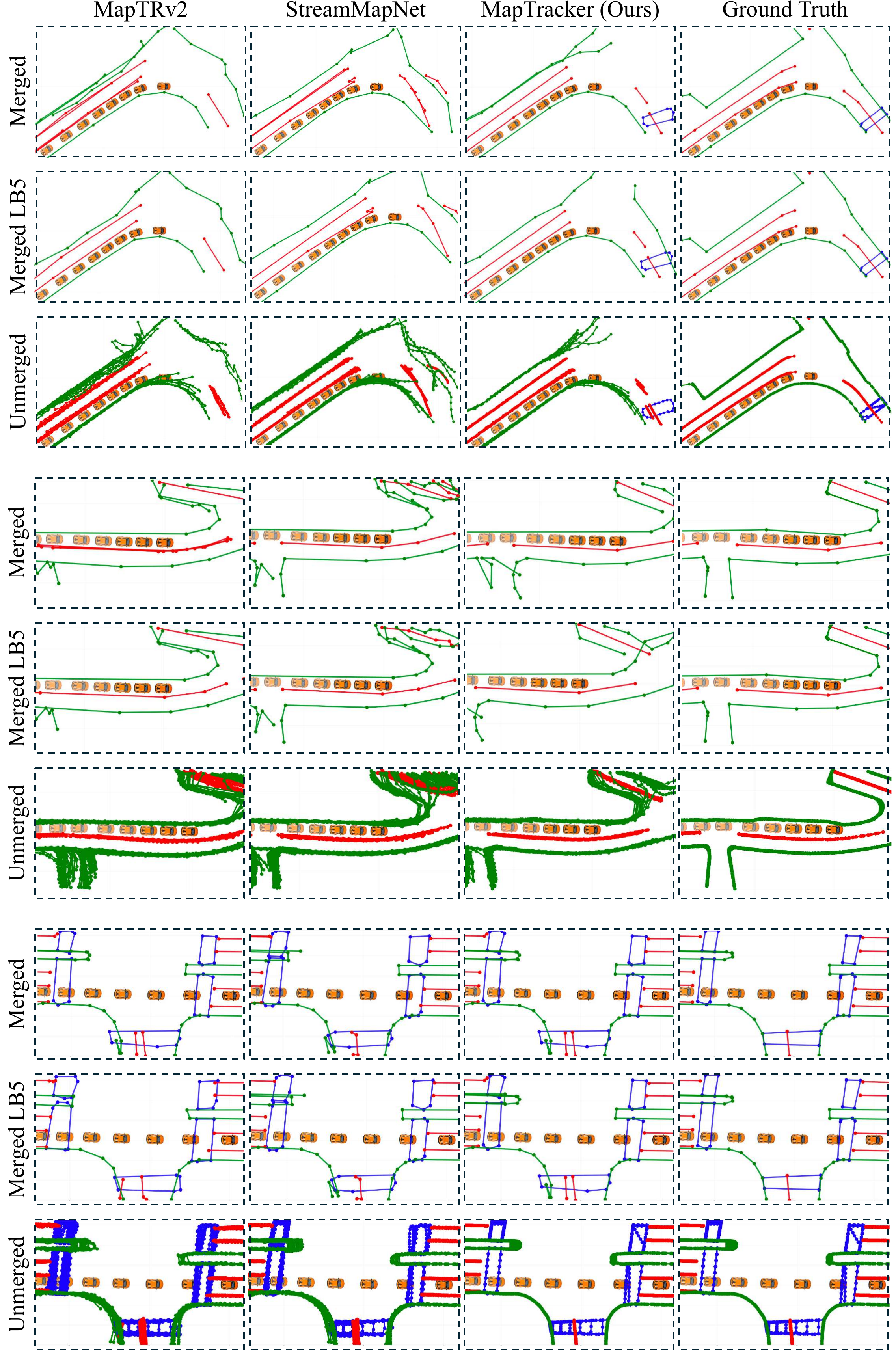}
\caption{Additional qualitative results on the nuScenes dataset.
}
\label{supp:qualitative:2}
\end{figure}

\begin{figure}[t]
\centering
\includegraphics[width=\textwidth]{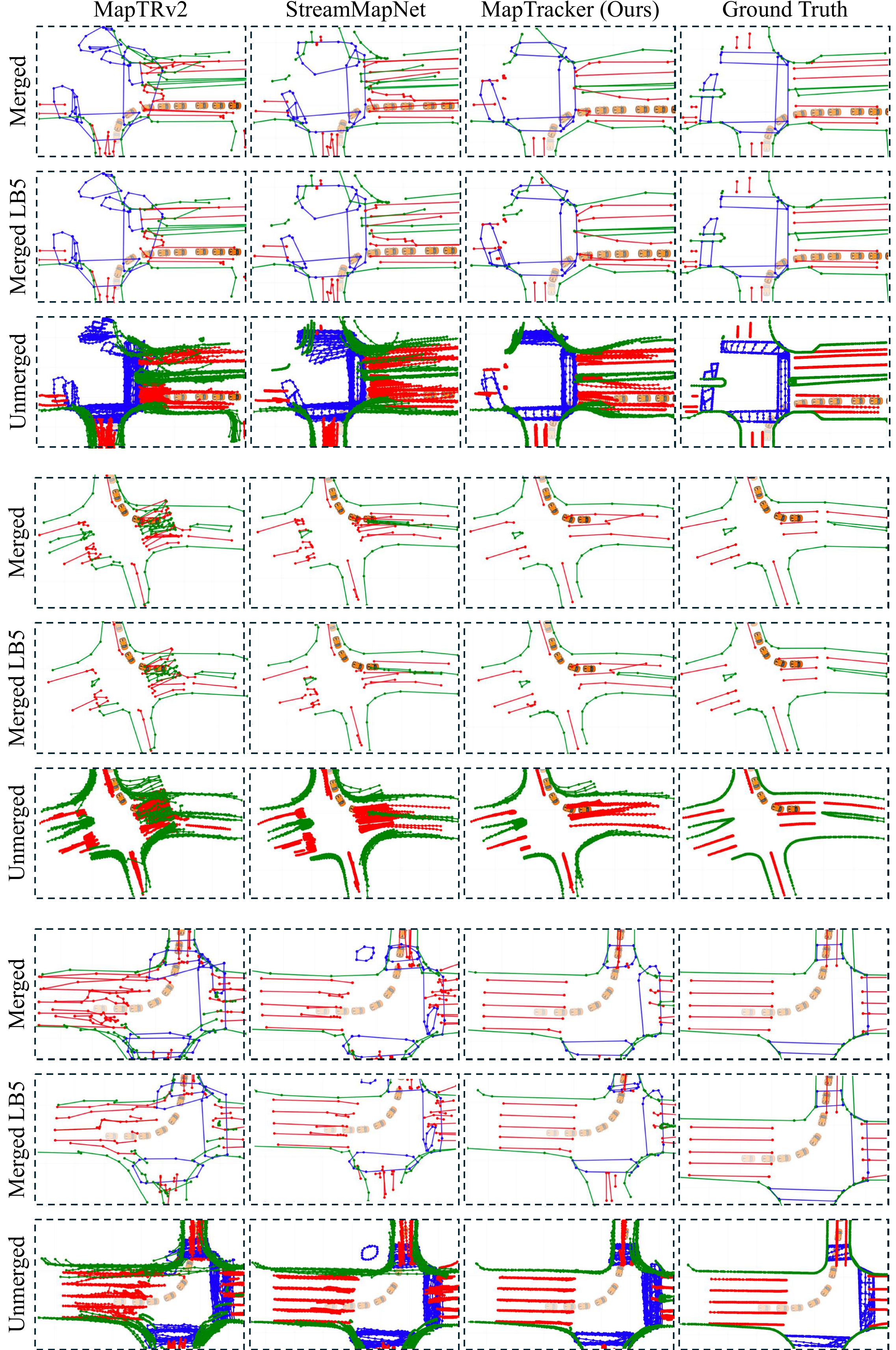}
\caption{Additional qualitative results on the nuScenes dataset.
}
\label{supp:qualitative:3}
\end{figure}

\begin{figure}[t]
\centering
\includegraphics[width=\textwidth]{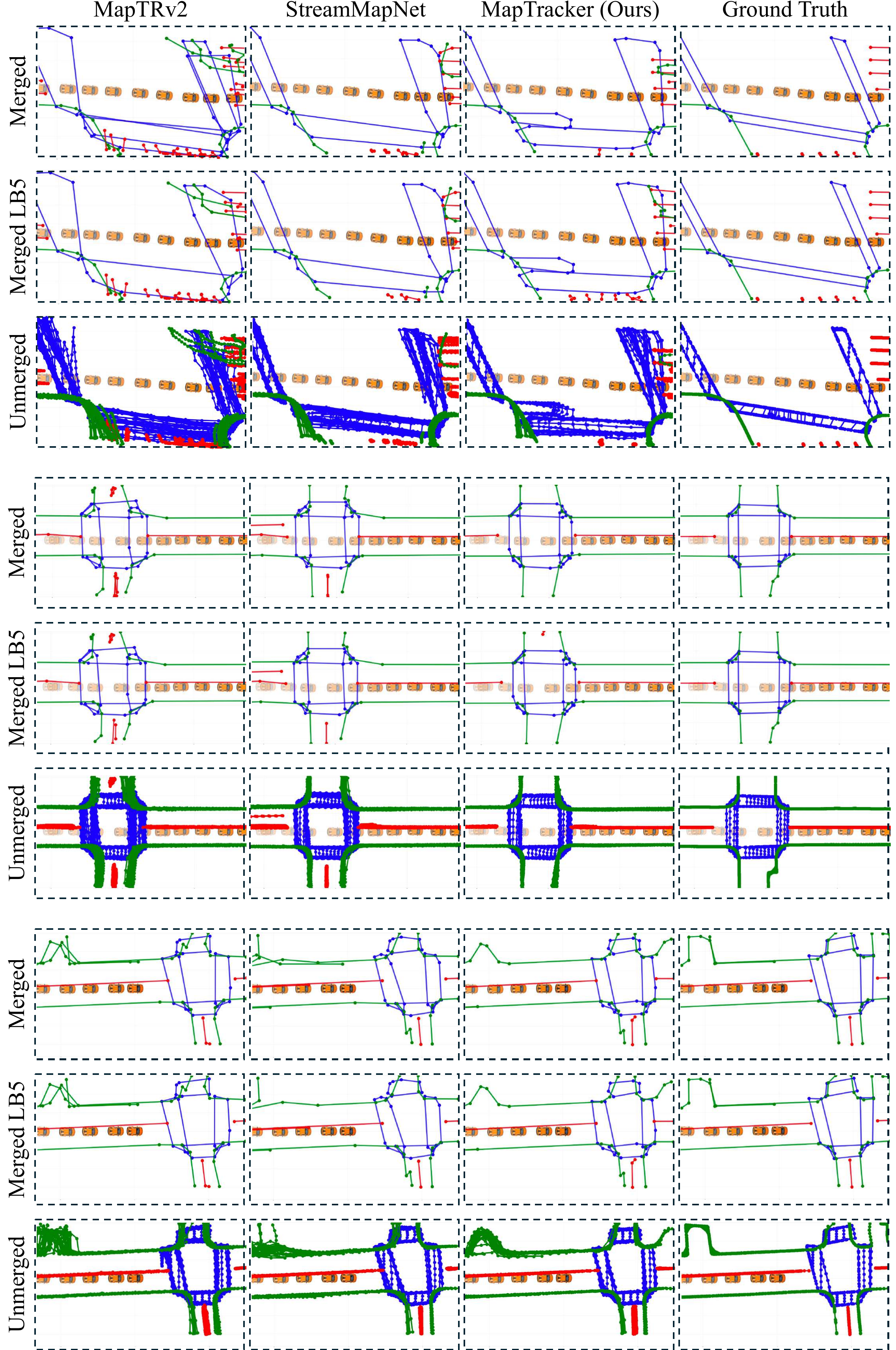}
\caption{Additional qualitative results on the Argoverse2 dataset.
}
\label{supp:qualitative:4}
\end{figure}

\begin{figure}[t]
\centering
\includegraphics[width=\textwidth]{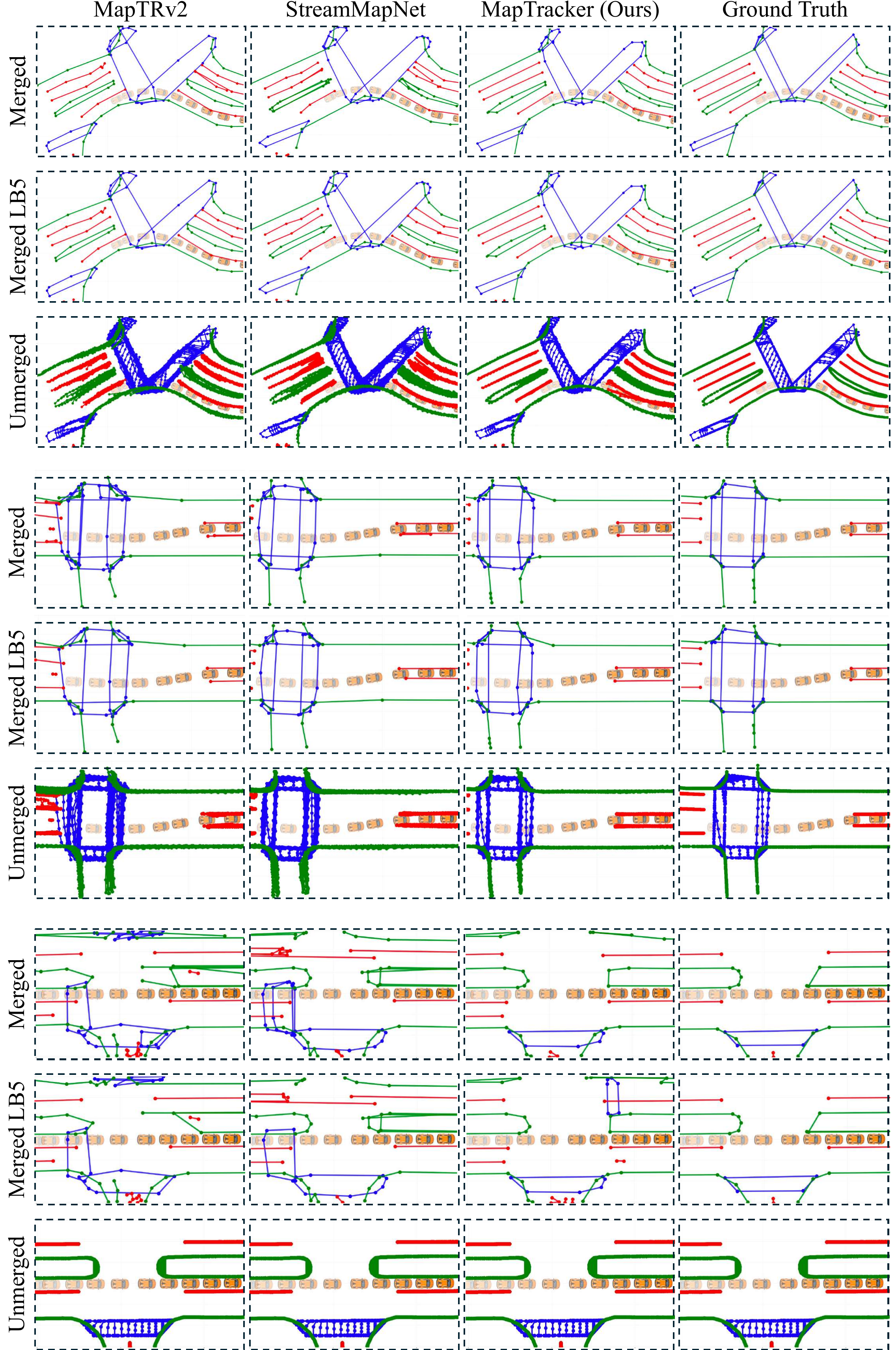}
\caption{Additional qualitative results on the Argoverse2 dataset.
}
\label{supp:qualitative:5}
\end{figure}

\begin{figure}[t]
\centering
\includegraphics[width=\textwidth]{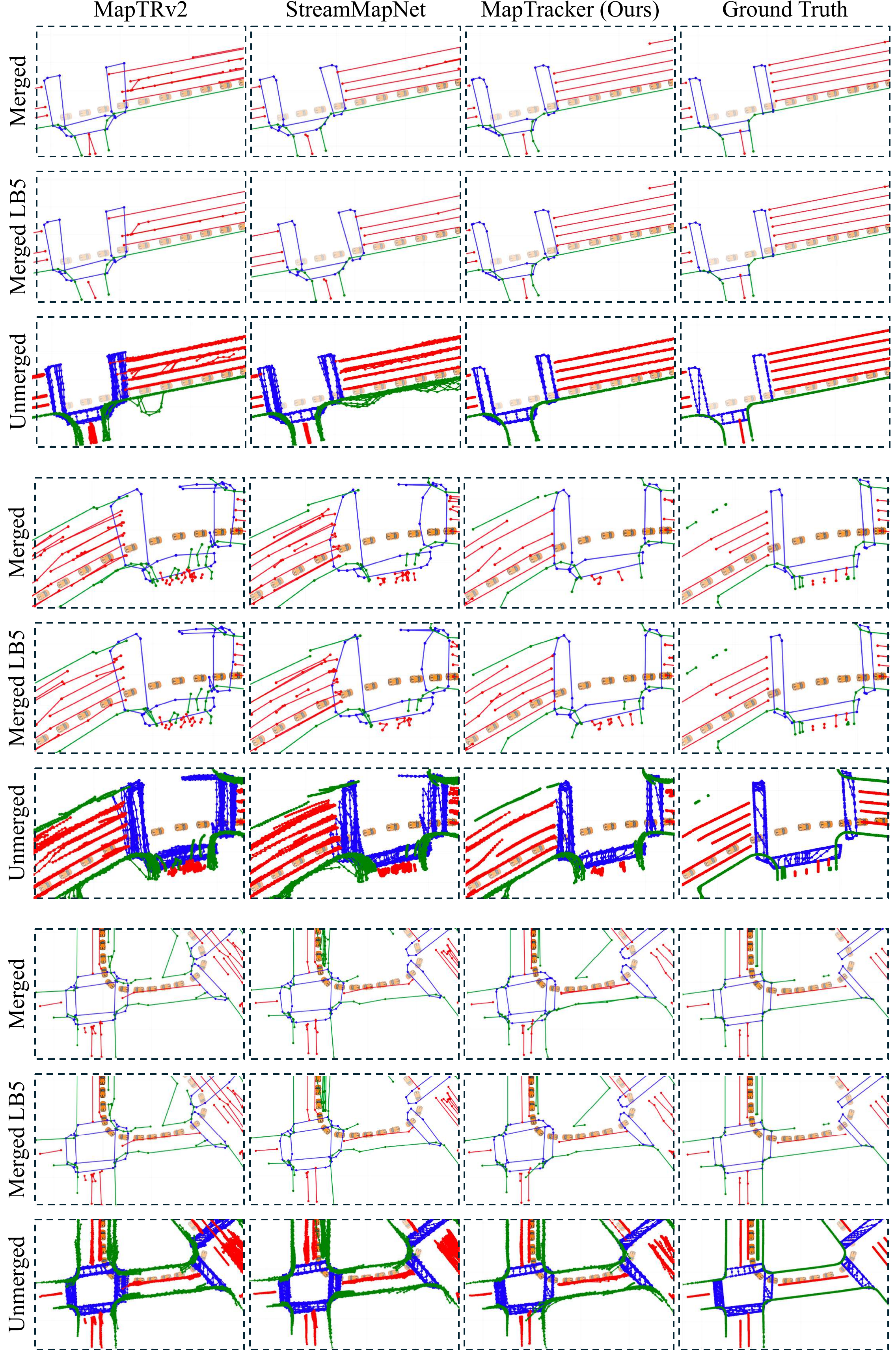}
\caption{Additional qualitative results on the Argoverse2 dataset.
}
\label{supp:qualitative:6}
\end{figure}